\documentclass[preprint,12pt]{elsarticle} % times

\biboptions{sort&compress}
\usepackage{lmodern}
\usepackage{graphicx}
\usepackage{multicol,multirow,booktabs}
\usepackage{amsmath,amssymb,amsthm}
\usepackage[utf8]{inputenc}
\usepackage[T1]{fontenc}
\usepackage[english]{babel}
\usepackage{tabularx}
\usepackage{algorithmicx}
\usepackage{algpseudocode}
\usepackage{subfigure}
\usepackage{url}

\newtheorem{remark}{Remark}
\newtheorem{algorithm}{Algorithm}

\usepackage{xcolor}

\theoremstyle{definition}

\usepackage{verbatim}

\DeclareMathOperator*{\maximise}{maximise}

\usepackage{array}
\newcolumntype{H}{>{\setbox0=\hbox\bgroup}c<{\egroup}@{}}

\definecolor{blue2b}{rgb}{0,0.1,0.3}
\definecolor{blue2}{rgb}{0,0.2,0.7}
\definecolor{red2}{rgb}{0.6,0.1,0.0}
\definecolor{green2}{rgb}{0.1,0.4,0.0}
\definecolor{yel2}{rgb}{0.3,0.2,0.0}
\definecolor{purple2}{rgb}{0.5,0.0,0.5}
\definecolor{blue3}{rgb}{0.65,0.85,1.0}
\definecolor{red3}{rgb}{1.0,0.7,0.5}
\definecolor{green3}{rgb}{0.8,1.0,0.7}
\definecolor{yel3}{rgb}{1.0,1.0,0.7}
\definecolor{grey3}{rgb}{0.95,0.95,0.95}
\definecolor{gray3}{rgb}{0.95,0.95,0.95}

\definecolor{grey1}{rgb}{0.30,0.30,0.30}
\definecolor{grey2}{rgb}{0.20,0.20,0.20}
\definecolor{grey0}{rgb}{0,0,0}

\begin{document}

%%%%%%%%%%%%%%%%%%%%%%%%%%%%%%%%%%%%%%%%%%%%%%%%%%%%%%%%%%%%%%%%%%%%%%%%%%%%%%%
%%%%%%%%%%%%%%%%%%%%%%%%%%%%%%%%%%%%%%%%%%%%%%%%%%%%%%%%%%%%%%%%%%%%%%%%%%%%%%%

\begin{frontmatter}

\title{Are Cluster Validity Measures (In)valid?}

\author[deakin,pas,cor1]{Marek Gagolewski}

\author[wut]{Maciej Bartoszuk}

\author[wut]{Anna Cena}

\address[deakin]{Deakin University, School of Information Technology, Geelong, VIC 3220, Australia}
\address[wut]{Warsaw University of Technology, Faculty of Mathematics and Information Science,\\%
ul.~Koszykowa 75, 00-662 Warsaw, Poland}
\address[pas]{Systems Research Institute, Polish Academy of Sciences\\%
ul.~Newelska 6, 01-447 Warsaw, Poland}

\cortext[cor1]{1) Corresponding author; email: m.gagolewski@deakin.edu.au}

\journal{Information Sciences}

%%%%%%%%%%%%%%%%%%%%%%%%%%%%%%%%%%%%%%%%%%%%%%%%%%%%%%%%%%%%%%%%%%%%%%%%%%%%%%%
%%%%%%%%%%%%%%%%%%%%%%%%%%%%%%%%%%%%%%%%%%%%%%%%%%%%%%%%%%%%%%%%%%%%%%%%%%%%%%%

\begin{abstract}
Internal cluster validity measures (such as the Caliński--Harabasz, Dunn,
or Davies--Bouldin indices) are frequently used for selecting the appropriate
number of partitions a dataset should be split into.
In this paper we consider what happens if we treat
such indices as objective functions in unsupervised learning activities.
Is the optimal grouping with regards to, say, the Silhouette index
really meaningful?
It turns out that many cluster (in)validity indices
promote clusterings that match expert knowledge quite poorly.
We also introduce a new, well-performing variant of the Dunn index that
is built upon OWA operators and the near-neighbour graph
so that subspaces of higher density,
regardless of their shapes, can be separated from each other better.

\end{abstract}
\end{frontmatter}

%%%%%%%%%%%%%%%%%%%%%%%%%%%%%%%%%%%%%%%%%%%%%%%%%%%%%%%%%%%%%%%%%%%%%%%%%%%%%%%
%%%%%%%%%%%%%%%%%%%%%%%%%%%%%%%%%%%%%%%%%%%%%%%%%%%%%%%%%%%%%%%%%%%%%%%%%%%%%%%
%%%%%%%%%%%%%%%%%%%%%%%%%%%%%%%%%%%%%%%%%%%%%%%%%%%%%%%%%%%%%%%%%%%%%%%%%%%%%%%

\section{Introduction}

An internal cluster validity index (CVI for short; see,
e.g., \cite{Milligan1985:psycho,Maulik2002:cvi_comp,
Halkidi2001:cluster_validity,ArbelaitzEtAl2013:extensive_CVI,
XU2020:external_synthetic}) is -- in theory -- a measure of
how well a given partitioning of a dataset
reflects the underlying structure of the modelled domain.
CVIs are frequently employed as tools for selecting the appropriate
number of clusters a dataset should be segmented into
\cite{Milligan1985:psycho}.
By re-applying some algorithm (e.g., $k$-means or spectral methods),
one can determine the splits into 2-, 3-, 4-, \dots,
disjoint and nonempty subsets, compute the corresponding CVIs,
and select the partition that maximises a chosen utility measure.

Here we shall focus on the other popular use case thereof.
Some practitioners utilise CVIs to compare the
outputs of different algorithms on the same dataset.
Is the partition that the average linkage method returned
better than that yielded by the DBSCAN algorithm (provided
that they are of equal cardinality)?
Similarly, researchers use CVIs for evaluating new clustering
algorithms the same way: a new method $X^{++}$ produces partitions that
have a higher average Caliński--Harabasz indices
(on some benchmark datasets) than procedures $X$, $Y$, and $Z$,
thus ``proving'' its superiority. However, we would like to call this
methodology into question, especially because CVIs in general
constitute an extremely diverse set of measures.

Thus, we shall be interested in determining which of the popular CVIs
are particularly suitable or unsuitable for judging the quality of different
partitions of the same cardinality. Does a high value of a CVI make sense at all?
Can it really be treated as an indicator of a useful clustering result?

To address these questions,
we shall find the partitions that yield the highest possible
index values, for a large number of datasets and CVIs.
In other words, we will treat each CVI as an objective
function to be maximised over the \textit{whole} space of \textit{all}
possible clusterings.

Our assumption here is that a cluster validity measure
can only be considered meaningful whenever it is maximised
at the partitions closely resembling the reference ones.
Otherwise stated, good CVIs should promote results
that agree with expert knowledge.

\begin{figure}[p!]
\centering
\includegraphics[width=0.49\linewidth]{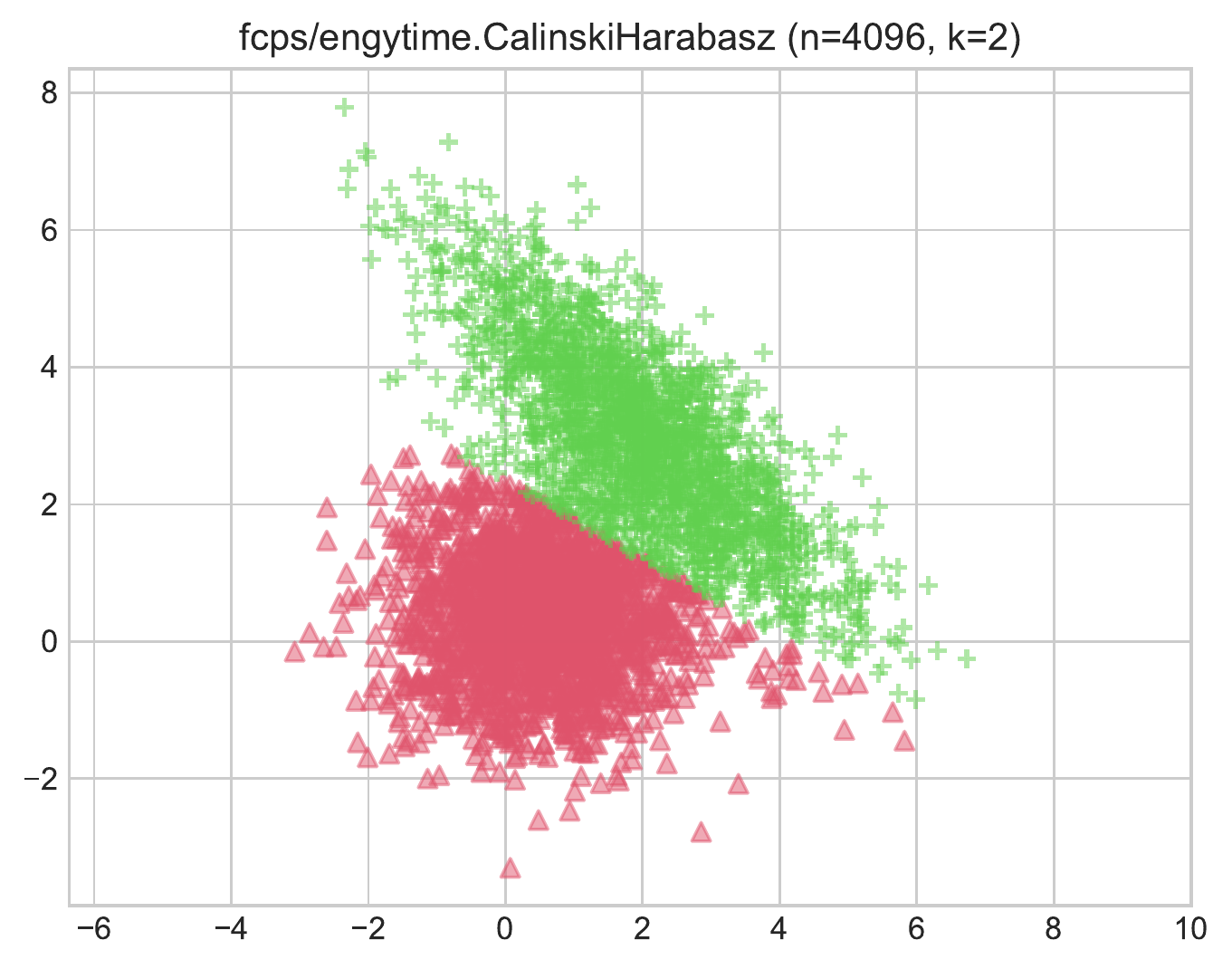}
\includegraphics[width=0.49\linewidth]{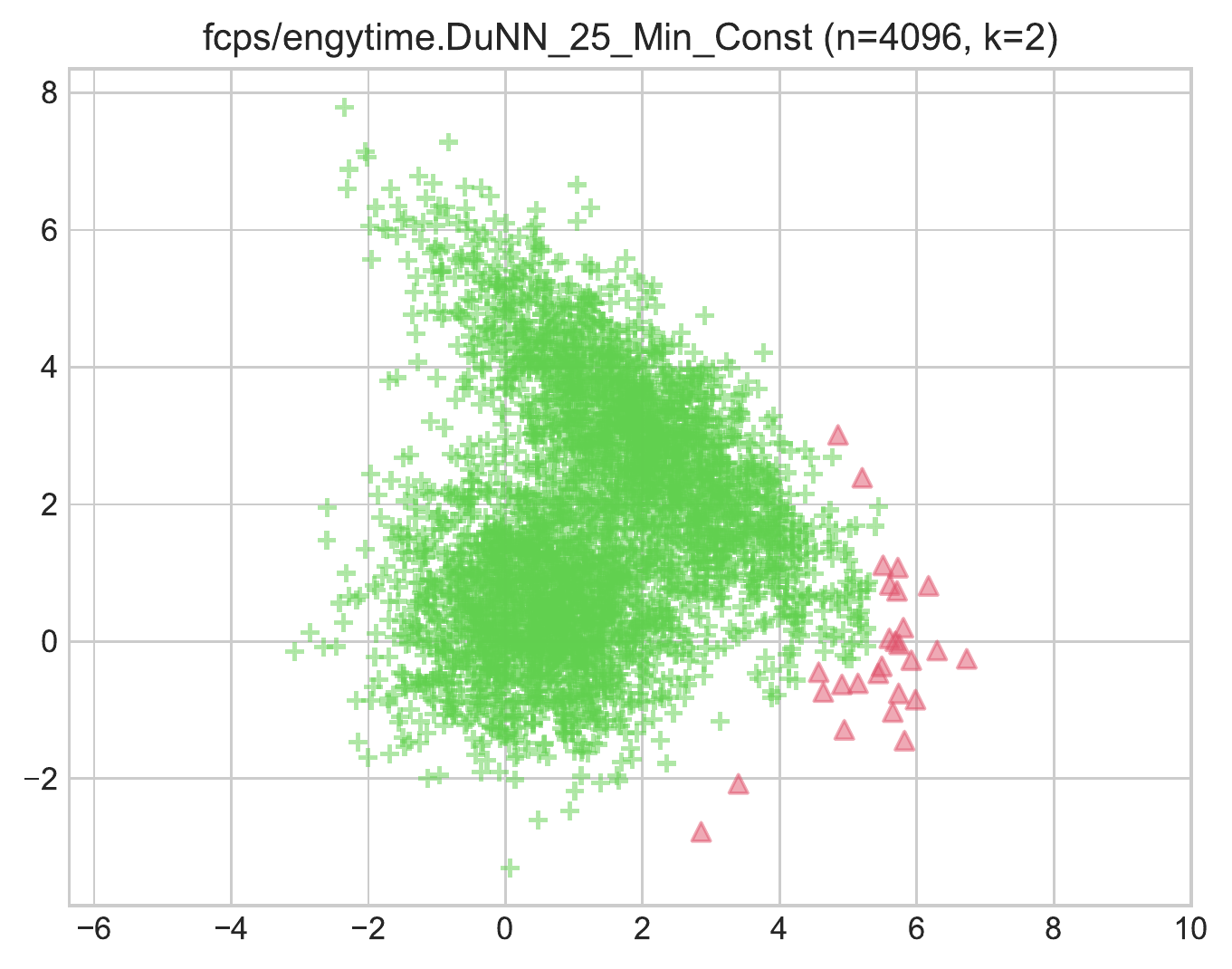}

\includegraphics[width=0.49\linewidth]{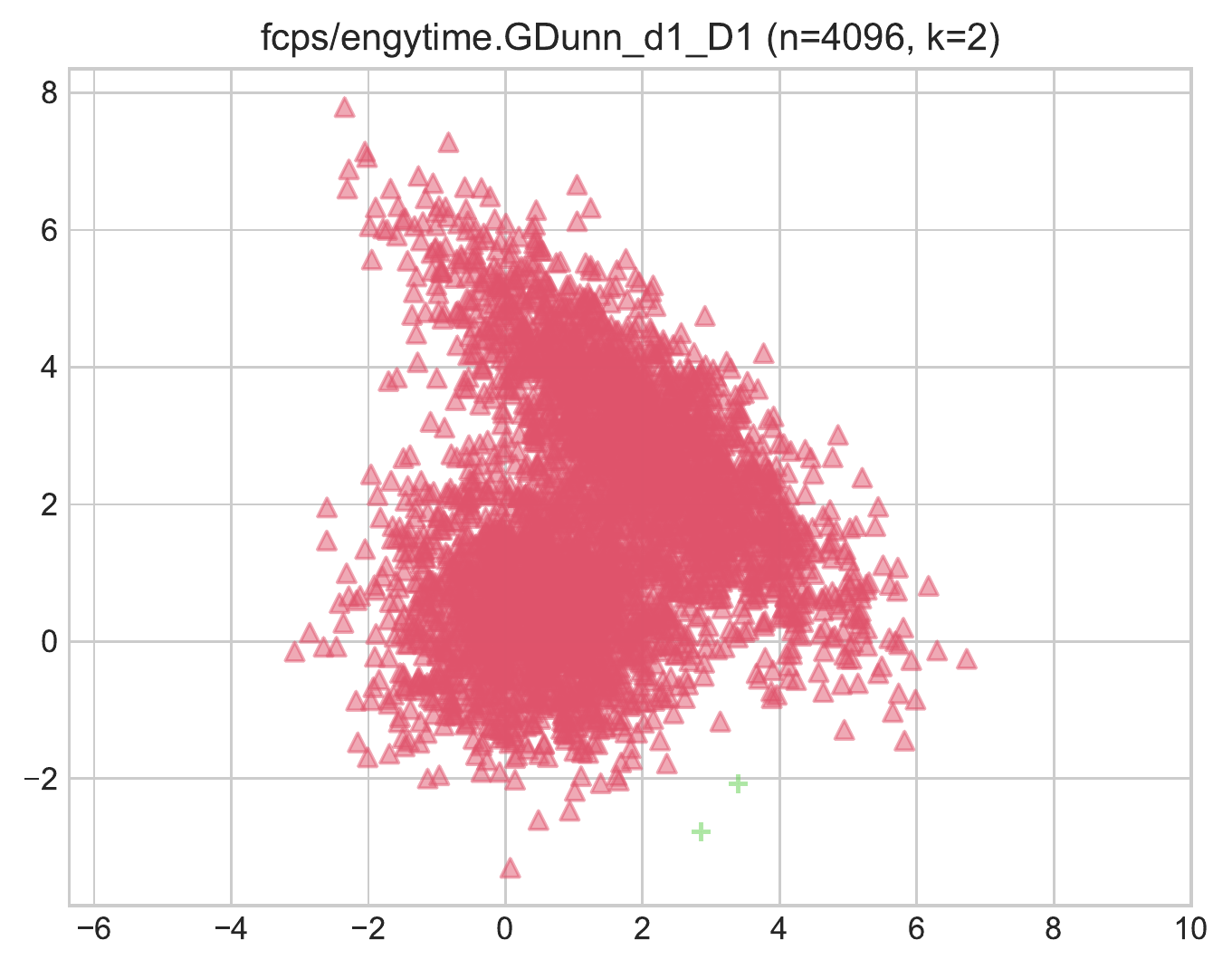}
\includegraphics[width=0.49\linewidth]{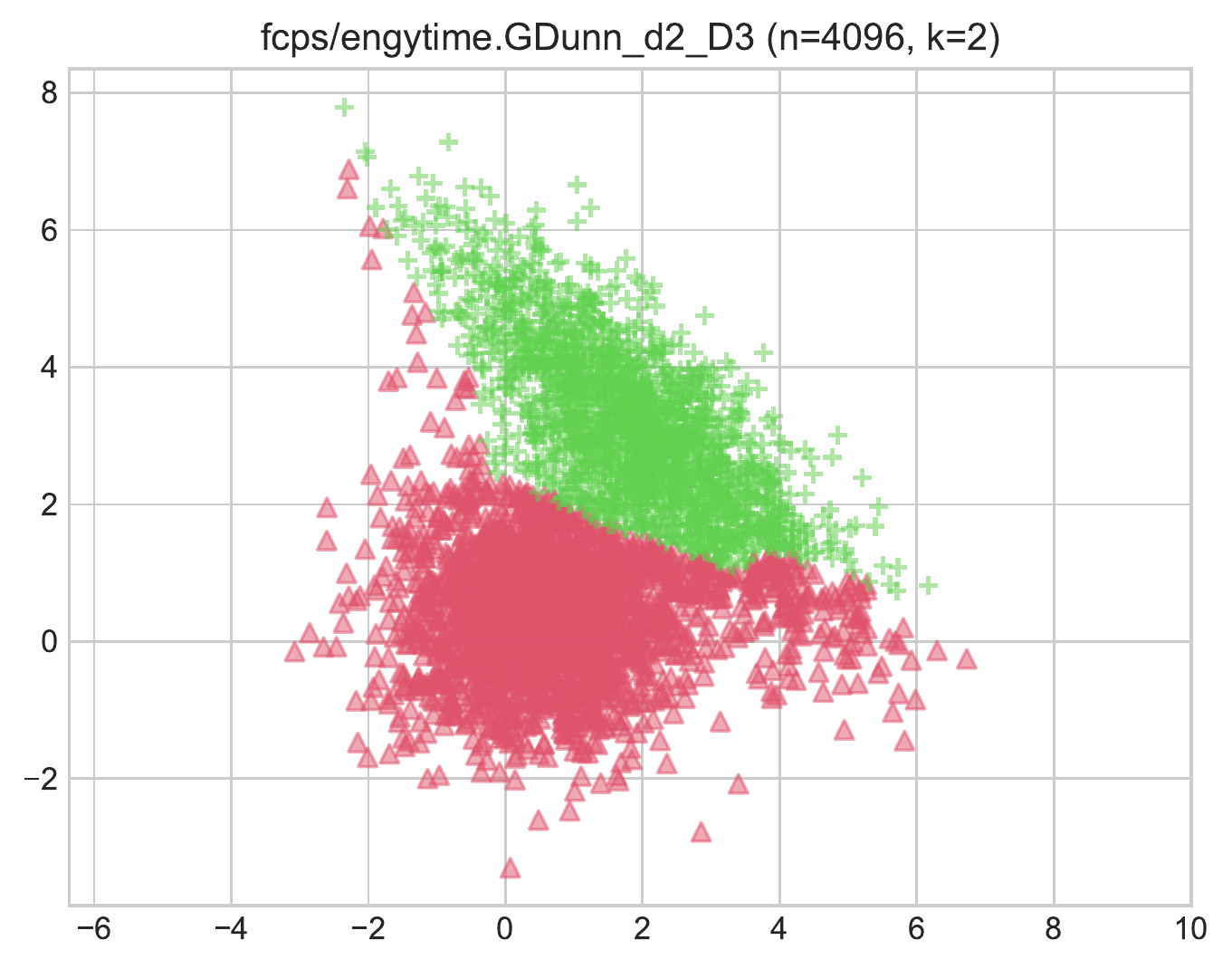}

\includegraphics[width=0.49\linewidth]{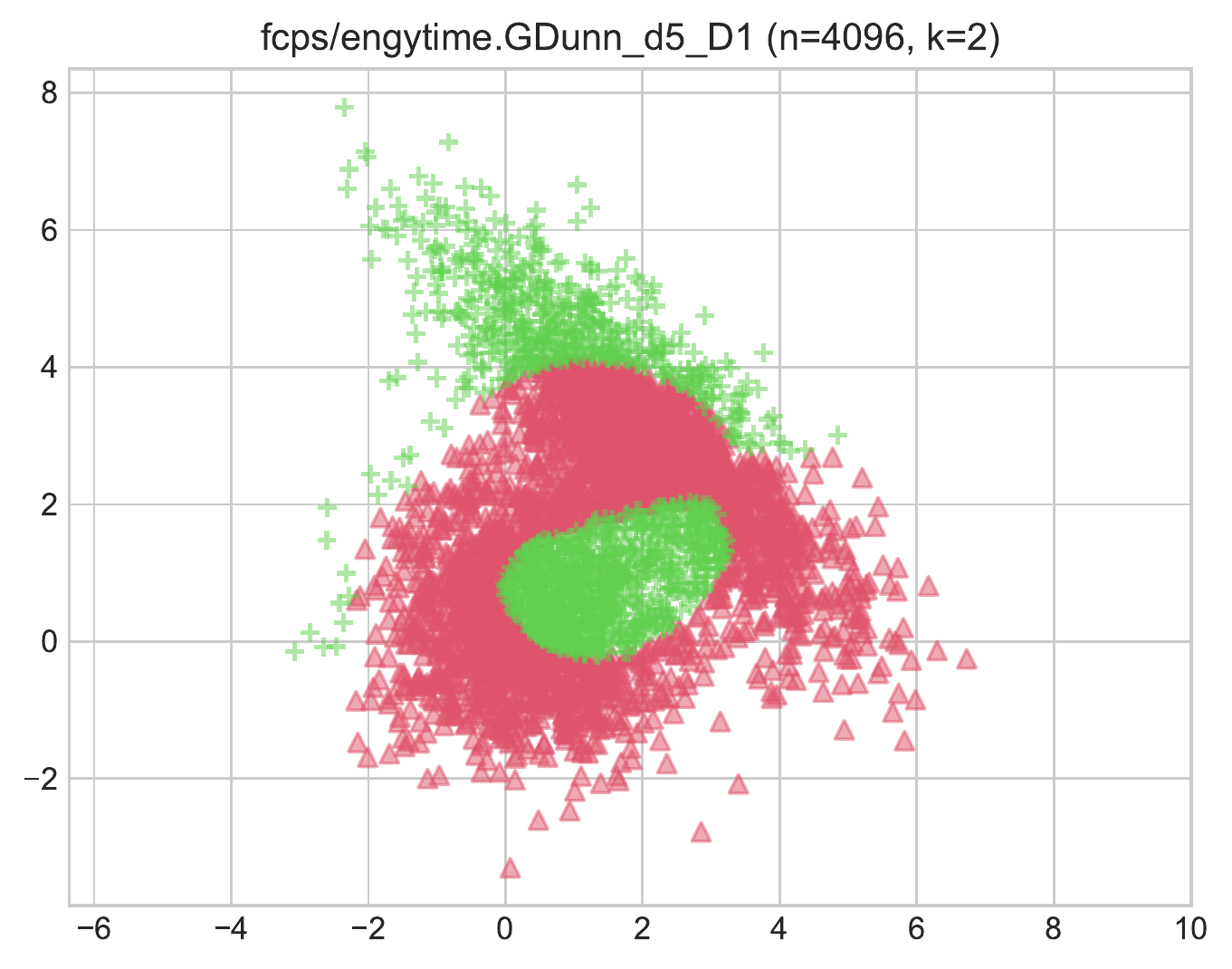}
\includegraphics[width=0.49\linewidth]{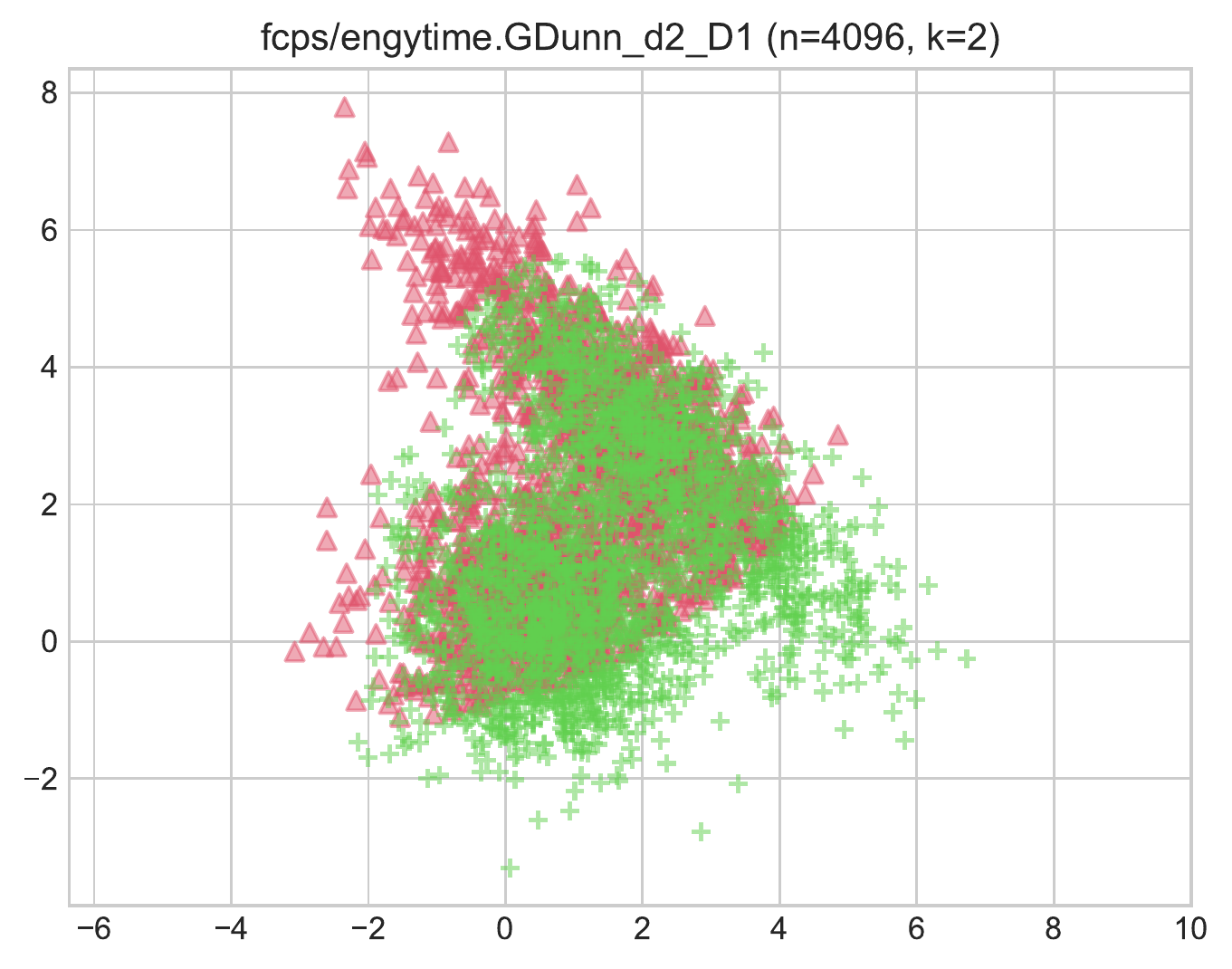}
\caption{\label{fig:maximisers_engytime}
\textit{fcps/engytime} dataset: Optimal clusters as seen by 6 different
cluster validity indices (see Section~\ref{sec:cvi} for more details).
The reference partitions consist of two (clearly visible if viewed in colour)
Gaussian blobs; the Caliński--Harabasz index (top-left subfigure) identifies them
quite correctly.
Some indices promote very peculiar, overlapping groupings (e.g.,
\textit{GDunn\_d2\_D1}),
other ones should rather be employed as outlier detectors
(like \textit{GDunn\_d1\_D1}).}
\end{figure}

\begin{figure}[p!]
\centering
\includegraphics[width=0.49\linewidth]{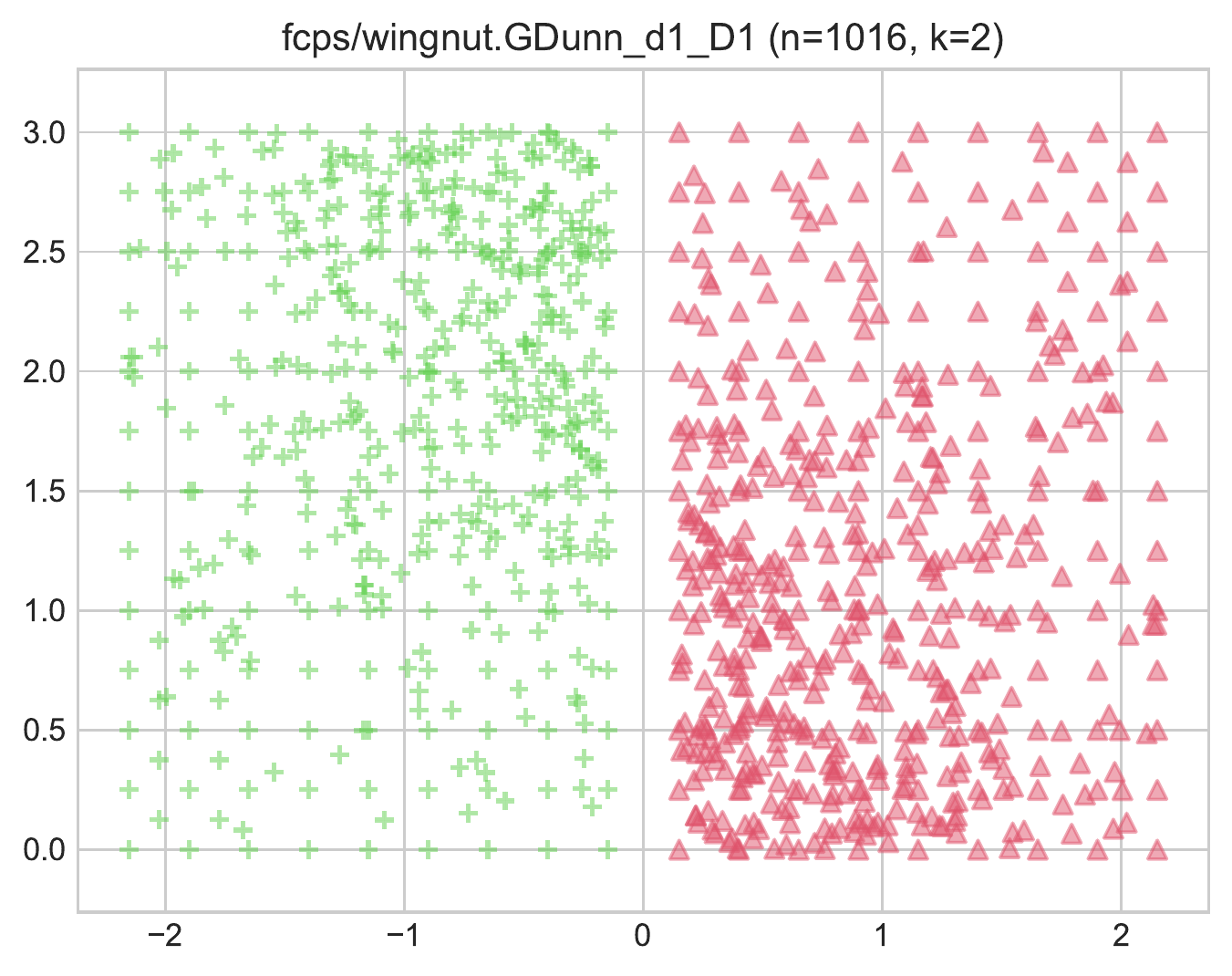}
\includegraphics[width=0.49\linewidth]{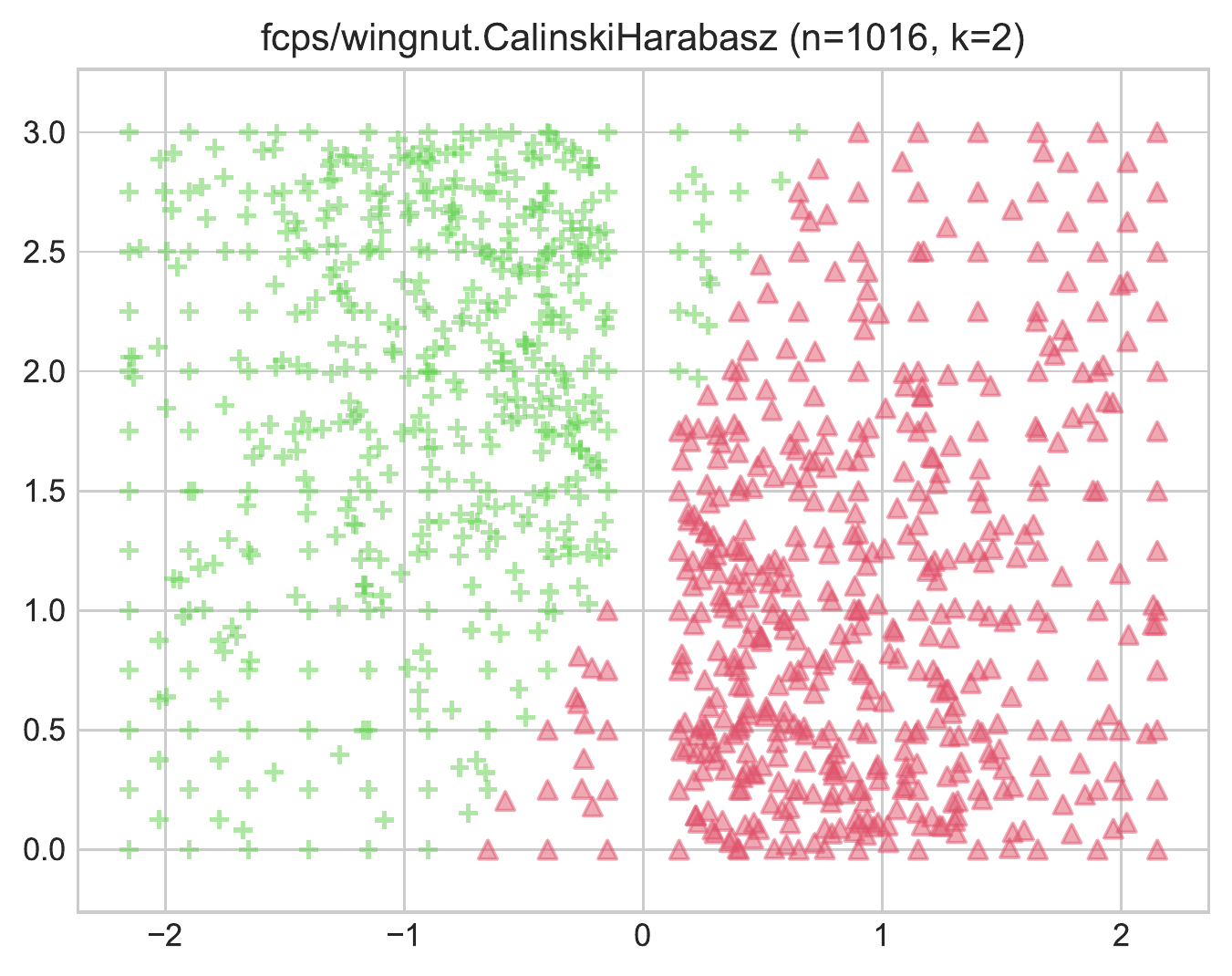}

\includegraphics[width=0.49\linewidth]{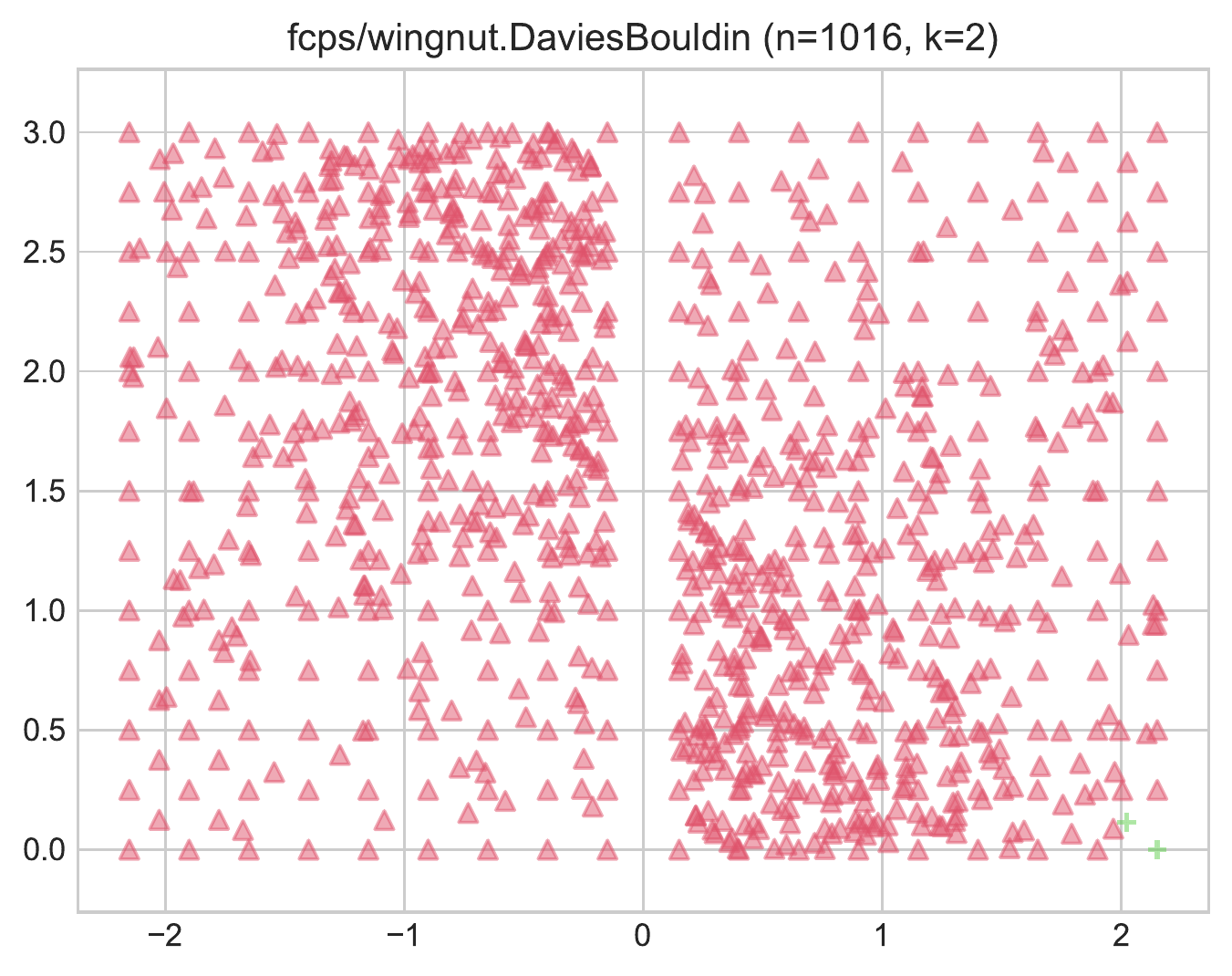}
\includegraphics[width=0.49\linewidth]{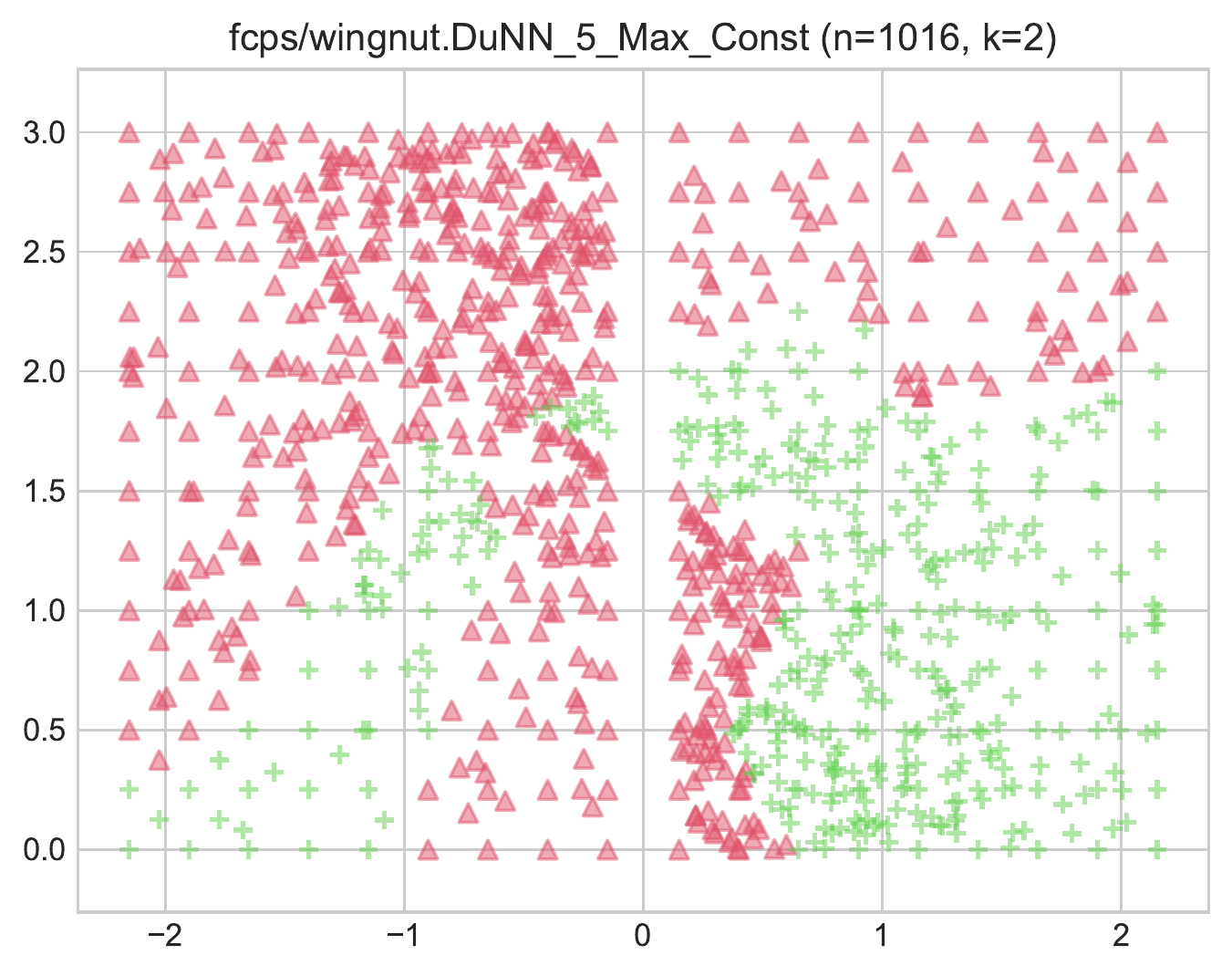}

\includegraphics[width=0.49\linewidth]{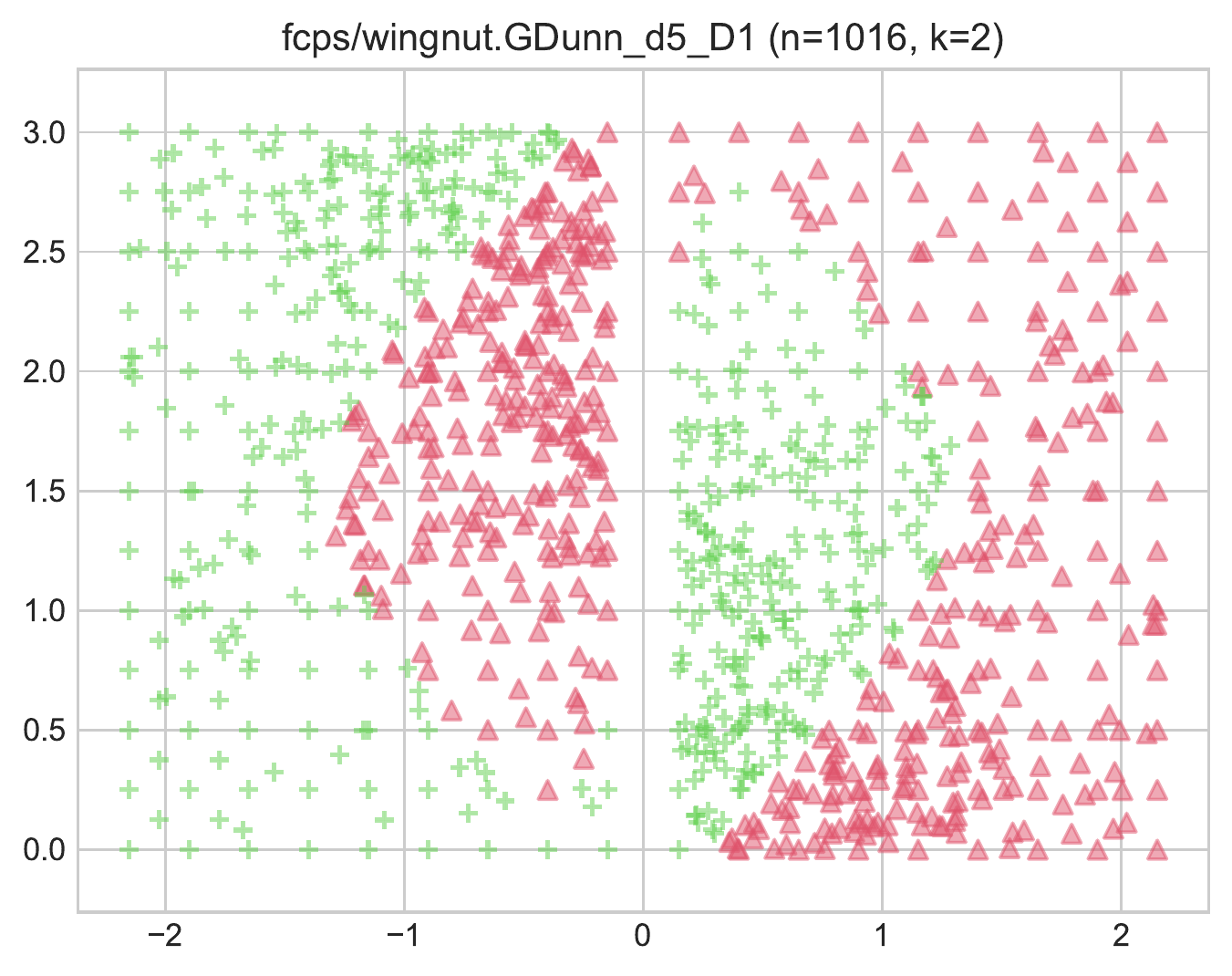}
\includegraphics[width=0.49\linewidth]{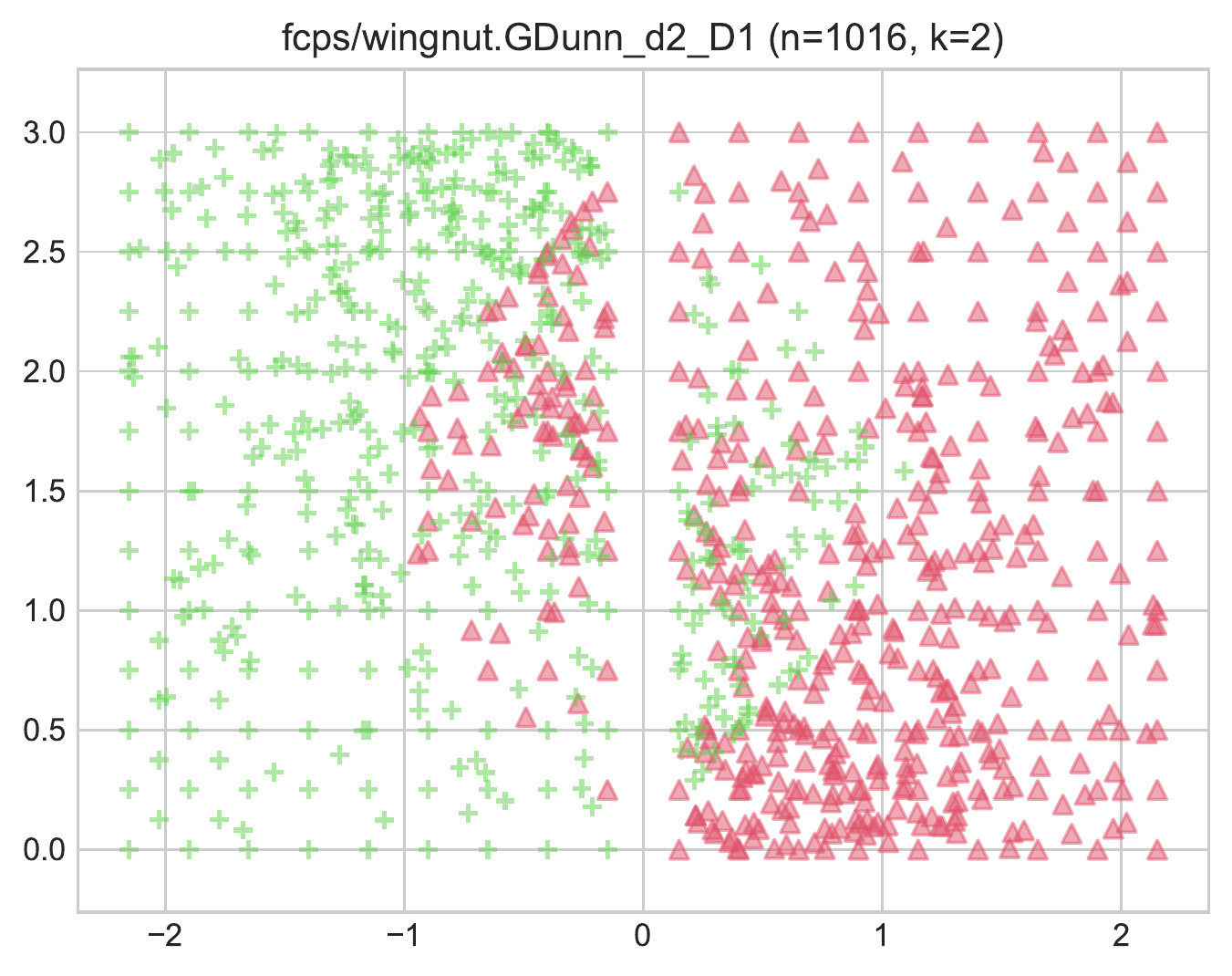}
\caption{\label{fig:maximisers_wingnut}
\textit{fcps/wingnut} dataset: Optimal clusters as seen by 6 different
cluster validity indices (discussed in Section~\ref{sec:cvi}).
The two point clouds are well-separable; \textit{GDunn\_d1\_D1}
(top-left subfigure) identified them correctly.
However, certain CVIs favour some rather unusual cluster shapes instead.}
\end{figure}

In the course of our study, which we of course detail
in the sequel, we have discovered
that this is often very much not the case -- see
Figures~\ref{fig:maximisers_engytime} and \ref{fig:maximisers_wingnut}
for two quite representative graphical examples.
It turns out that some CVIs promote highly overlapping groupings
while other ones work better as outlier detectors.
One should thus not uncritically believe that a high value of, e.g.,
a generalised Dunn index  \textit{GDunn\_d2\_D1}
(see Section~\ref{sec:cvi}) is better than a lower one;
at the bottom-right subfigures we see that this index
promotes some rather random partitions as ``best''.

\clearpage

\bigskip
Let us note that papers introducing new CVIs are plentiful and
new ones are being published on a regular basis, see, e.g.,
\cite{LIU2021579,LIANG2020106583} for some recent examples.
More often than not, such indices are designed to be ``the best'' for a given
particular situation and/or they aim to ``eliminate'' certain deficiencies
with the previous measures.
Because of this, the number of CVIs to choose from can be overwhelming,
see Section~\ref{sec:cvi} and, e.g., \cite{XU2020:external_synthetic,VijKhandnor2016:CVIsummary},
for up-to-date overviews.

This is why comprehensive, in-depth comparative studies are so important.
There are quite a few interesting surveys of various properties
of cluster validity measures in different contexts and settings, but
we have only found \cite{6170593} somewhat methodologically relevant
to the task at hand.
There, the performance of 8 CVIs was studied but the authors' focus
seems slightly shifted towards the analysis of the stability of the solutions
generated by what they call
differential-evolution--particle-swarm-optimisation algorithms,
which of course do not necessarily guarantee finding
patterns that are optimal in the eye of a specific cluster validity measure.

Further, in \cite{ArbelaitzEtAl2013:extensive_CVI}
and  \cite{Maulik2002:cvi_comp},
the authors have shown that most of the CVIs work well with spherical
clusters but fail in other types of data. A CVI
may fail to assign the highest evaluation to the partition that
fits the data best, e.g., the reference one.
The conclusion is that there is no single internal cluster validation
index that outperforms the other indices everywhere.
Similar conclusions were reached in \cite{BrunETAL2007:modelcvi},
where a model-based study of the correlations of CVIs with carefully
chosen error rates across the outputs of different clustering
algorithms was conveyed. For more studies in similar spirit, see
\cite{Dimitriadou1999:more_insights,
KimRamakrishna2005:CVIassessment,Milligan1985:psycho,rs8040295}.
Note again that our task is to find the optimal partition under
the guidance of a given CVI (amongst the set of all partitions),
and not to assess a fixed set of particular clusterings.

\bigskip
We are well aware of the fact that  the sheer act of optimising of CVIs
is, overall, not a new idea.
For instance, the within-cluster sum of squares (WCSS),
which is the basis for the Caliński--Harabasz index %and Ball--Hall indices
is used as the objective function in the $k$-means \cite{lloyd} algorithm.

Other algorithms employing some goodness-of-split measures
include, e.g., fuzzy $c$-means \cite{Bezdek} that features
a smoothened variant of WCSS,
ITM (Infor\-mation-Theoretic-MST; \cite{itm})
which applies a divisive scheme over an
Euclidean minimum spanning tree to optimise an information-theoretic criterion,
finding Gaussian mixtures via expectation-maximisation,
and the generalisation of the Ward linkage in the form of the
Lance--Williams formulae~\cite{LanceWilliams1967:hierarchicalformula}.

Nevertheless, as $n$ data items enjoy $\Omega(k^n)$
possible $k$-partitionings (Stirling number of the second kind),
the problem of optimising a general CVI over the whole search space is
very difficult computationally (provably hard in the case of the
said WCSS, see \cite{kmeanshard,kmeanseuclidhard}, amongst many others).

Because of this, many algorithms which are defined
as minimisers/maxi\-misers of some CVI, use a variety of
simplifications, approximations, or heuristics,
e.g., they optimise the objective over reduced search spaces or
apply some greedy strategies.
For instance, to find a partition optimising
the aforementioned WCSS,
Ward in \cite{Ward1963:hier} suggested to build a hierarchy of
clusters in an agglomerative way,
and Edwards and Sforza in \cite{EdwardsSforza1965:divisive}
as well as Caliński and Harabasz in \cite{CalinskiHarabasz1974:index}
proposed to employ some divisive schemes.
However, such heuristics are usually limited to specific CVIs;
in this paper we would like to go far beyond that.

Also notice that in the literature we may of course find numerous
techniques constructed as a combination of different metaheuristics-based
optimisation procedures and chosen cluster validity
measures playing the role of objective functions, see, e.g.,
\cite{8279595,KUO20211,Qadura2020:nn_ev_cvi,
ZhuEtAl2020:evolution_cvi,NANDA20141,DHAL2021106814}.
Their respective authors often claim that this way they create
``new'' clustering algorithms
(e.g., optimise WCSS using particle swarms vs~by means of
differential evolution), but it is semantically  not quite appropriate.
Rather, they should be thought of as ways to test the
performance of the optimisation algorithms themselves
(with CVIs computed over particular datasets serving as benchmark objectives
as in \cite{benchmarkoptim}).

\bigskip
Due to the intrinsic difficulty of optimising CVIs,
it is no wonder that such a comprehensive
study as the current one has not been performed
yet. Interestingly, it will turn out that the methodology we have employed,
despite its still being  based on some approximations, is sufficient
for achieving our goal.

Also it is worth mentioning that, up to date, the variety of studies of the
various aspects of CVIs (which we shall review in the next section),
was quite limited because of the lack of a larger, standardised benchmark set
batteries. Most studies considered few datasets, either synthetic, or
inherently difficult to partition (such as the UCI \cite{uci} datasets).
Luckily, thanks to the recent notable efforts
by the authors of \cite{fcps,GravesPedrycz2010:kernelfuzzyclust,kmsix}
and our new battery that aggregates and extends them
\cite{clustering_benchmarks_v1},
studies such as this one  finally become possible.
Furthermore, we propose a unique  approach where
we account for the fact that a dataset can exhibit multiple
equally valid clusterings (as discussed also in \cite{multiple_clusterings}).

%%%%%%%%%%%%%%%%%%%%%%%%%%%%%%%%%%%%%%%%%%%%%%%%%%%%%%%%%%%%%%%%%%%%%%%%%%%%%%%%

\bigskip
Let us cast a glance at the structure of this paper.
In Section~\ref{sec:cvi} we review some of the most notable cluster
validity indices. Furthermore,
we propose a new, wide class of CVIs
generalising the Dunn index which is based on the notion of ordered weighted
averaging (OWA) operators and near-neighbour (NN) graphs.
In Section~\ref{sec:method} we describe the methodology we have applied
in order to answer our main research question, including the description of
the benchmark datasets used,
the approach to identify the partitions that are optimal
from the perspective of a given cluster validity measure,
and ways to determine the extent to which they agree with expert knowledge.
In Section~\ref{sec:experiment} we present the results of our
empirical study, e.g., explore the relationship between cluster compactness
or separability and what is considered a good clustering by experts.
Also, we perform a cluster analysis of clustering
algorithms to determine which methods are most similar to each other.
We conclude the paper in Section~\ref{sec:conclusion}.

%%%%%%%%%%%%%%%%%%%%%%%%%%%%%%%%%%%%%%%%%%%%%%%%%%%%%%%%%%%%%%%%%%%%%%%%%%%%%%%

\section{Cluster validity indices}\label{sec:cvi}

Let $\mathbf{X}\in\mathbb{R}^{n\times d}$ denote the input dataset
comprised of $n$ points in a $d$-dimensional Euclidean space,
with $\mathbf{x}_i = (x_{i,1},\dots,x_{i,d})$ denoting
the $i$-th point, $i\in[1:n]=\{1,2,\dots,n\}$.

We shall be looking for a partition of $\mathbf{X}$ into $k\ge 2$ nonempty,
mutually disjoint clusters, with $k$ fixed in advance.
Note that a $k$-partition $\{X_1,\dots,X_k\}$ of a set
$\{\mathbf{x}_1, \dots, \mathbf{x}_n\}$
can be encoded by means of a surjection
$C: [1:n]\stackrel{\text{onto}}{\to}[1:k]$,
where $C(i)\in[1:k]$ gives the cluster number of the $i$-th point.
Let us denote the set of all such possible mappings with $\mathcal{C}_k$.

\bigskip
For the sake of clarity and simplicity, we will only
be focused on cluster validity indices based
on Euclidean distances between all pairs of points,
$\|\mathbf{x}_i-\mathbf{x}_j\|$,
or the input points and some other pivots,
such as their corresponding cluster centroids,
$\|\mathbf{x}_i-\boldsymbol\mu_j\|$, where
$\mu_{j,l} = \frac{1}{|X_j|} \sum_{\mathbf{x}_i\in X_j} x_{i,l}$.
The fixation of the distance metric is not at all
restrictive, as various
transformations can be applied onto $\mathbf{X}$ at the data pre-processing
stage, including variable selection, standardisation, outlier removal,
feature engineering (by means of spectral/kernel-based methods), etc.,
see \cite{DINH2021418,MISHRA2021,DU2021375}, amongst others.

\bigskip
We shall consider 52 different internal cluster validity indices
like $I:\mathcal{C}_k\to\mathbb{R}$.
Apart from the most popular, classical CVIs,
we also bring forth  our own proposal.

\subsection{CVIs based on cluster centroids}

\paragraph{1,2) BallHall, CalińskiHarabasz}
Let $\boldsymbol\mu$ denote the centroid of the whole $\mathbf{X}$.
The two following indices are based on within-cluster sum of squares (WCSS),
which itself can be rewritten in terms of the squared Euclidean
distances between the points and their respective centroids.

The Ball--Hall index  \cite{BallHall1965:isodata}
is the WCSS weighted by the cluster cardinality:
\begin{equation}
\mathrm{BallHall}(C) =
-
\sum_{i=1}^n \frac{1}{|X_{C(i)}|}  \| \mathbf{x}_i - \boldsymbol\mu_{C(i)} \|^2.
\end{equation}
Note the minus that accounts for the fact that in Section~\ref{sec:optim}
we want all the indexes be maximised.

Then the Caliński--Harabasz index  \cite[Eq.~(3)]{CalinskiHarabasz1974:index}
(``variance ratio criterion'')
is given by:
\begin{equation}
\mathrm{CalińskiHarabasz}(C) =
\frac{n-k}{k-1}
\frac{
\sum_{i=1}^n   \| \boldsymbol\mu - \boldsymbol\mu_{C(i)} \|^2
}{
\sum_{i=1}^n   \| \mathbf{x}_i - \boldsymbol\mu_{C(i)} \|^2
}.
\end{equation}
It may be shown that the task of minimising the (unweighted) WCSS
is equivalent to maximising the Caliński--Harabasz index.
Hence, this index is precisely the objective function
in $k$-means \cite{lloyd} and the algorithms by Ward,
Edwards and  Cavalli-Sforza, etc., see \cite{Ward1963:hier,
EdwardsSforza1965:divisive, CalinskiHarabasz1974:index}.

\paragraph{3) DaviesBouldin}
The Davies--Bouldin \cite[Def.~5]{DaviesBouldin1979:index} index
also refers to the notion of cluster centroids.
It is given as the average similarity between each cluster
and its most similar counterpart
(note the minus sign again):
\begin{equation}
\mathrm{DaviesBouldin}(C) =
-\frac{1}{k}
\sum_{i=1}^k\left(
\max_{j\neq i}
\frac{s_i+s_j}{m_{i,j}}
\right),
\end{equation}
where $s_i$ is the dispersion of the $i$-th cluster:
if $|X_i|>1$, it is given by
$s_i=\frac{1}{|X_i|}\sum_{\mathbf{x}_u\in X_i} \|\mathbf{x}_u-\boldsymbol\mu_i\|$
and otherwise we set $s_i=\infty$.
Furthermore,
$m_{i,j}$ is the intra-cluster distance,
$m_{i,j}=\|\boldsymbol\mu_i-\boldsymbol\mu_j\|$.
In \cite{DaviesBouldin1979:index}, other choices of $s_i$ and $m_{i,j}$
are also suggested; here, we choose the most popular setting
(used, e.g., in \cite{ArbelaitzEtAl2013:extensive_CVI}).

\subsection{Silhouettes}

\paragraph{4, 5) Silhouette, SilhouetteW}
In \cite[Sec.~2]{Rousseeuw1987:silhouettes},
Rousseeuw proposes the notion of a silhouette as a graphical
aid in cluster analysis.

Denote the average dissimilarity between the $i$-th point
and all other points in its own cluster with:
\begin{equation}
a_i = \frac{1}{|X_{C(i)}|-1} \sum_{\mathbf{x}_u\in X_{C(i)}} \| \mathbf{x}_i-\mathbf{x}_u \|
\end{equation}
and the average dissimilarity between the $i$-th point
and all other entities in the ``closest'' cluster with:
\begin{equation}
b_i = \min_{j\neq C(i)} \left(
\frac{1}{|X_j|} \sum_{\mathbf{x}_v\in X_{j}} \| \mathbf{x}_i-\mathbf{x}_v \|
\right).
\end{equation}
Then the \textit{Silhouette} index is defined
as the average silhouette score:
\begin{equation}
\mathrm{Silhouette}(C) = \frac{1}{n}\sum_{i=1}^n \frac{b_i-a_i}{\max\{ a_i, b_i \}},
\end{equation}
with convention $\pm\infty/\infty=0$.

The same paper also defines what we  call here the \textit{SilhouetteW} index,
being the mean of the cluster average silhouette widths: %(widths=true)
\begin{equation}
\mathrm{SilhouetteW}(C) = \frac{1}{k-s}
\sum_{i=1}^n
\frac{1}{|X_{C(i)}|}
\frac{b_i-a_i}{\max\{ a_i, b_i \}},
\end{equation}
where $s$ is the number of singletons.
Note that \textit{SilhouetteW}, just like \textit{BallHall},
employs weighting by cluster cardinalities.

\subsection{Generalised Dunn indices}

\paragraph{6--20) GDunn\_dX\_DY}
In \cite[Eq.~(3)]{Dunn1974:index}, Dunn proposed an index defined as
the ratio between the smallest between-cluster distance  and
the largest cluster diameter.
It has been generalised by Bezdek and Pal in \cite{BezdekPal1998:gdunn} as:
\begin{equation}
\mathrm{GDunn}(C)=
\frac{
\min_{i\neq j} d\left( X_i, X_j \right)
}{
\max_{i} D\left( X_i \right)
}.
\end{equation}
The numerator measures the between-cluster separation whilst
the denominator quantifies the cluster compactness.

Function $d$ was assumed in \cite{BezdekPal1998:gdunn} one of:
\begin{itemize}
\item $d_1(X_i, X_j)=\mathrm{Min}\left(
\left\{ \|\mathbf{x}_{u}-\mathbf{x}_{v}\|: \mathbf{x}_{u}\in X_i, \mathbf{x}_{v}\in X_j\right\}
\right)$,
\item $d_2(X_i, X_j)=\mathrm{Max}\left(
\left\{ \|\mathbf{x}_{u}-\mathbf{x}_{v}\|: \mathbf{x}_{u}\in X_i, \mathbf{x}_{v}\in X_j \right\}
\right)$,
\item $d_3(X_i, X_j)=\mathrm{Mean}\left(
\left\{ \|\mathbf{x}_{u}-\mathbf{x}_{v}\|: \mathbf{x}_{u}\in X_i, \mathbf{x}_{v}\in X_j \right\}
\right)$,
\item $d_4(X_i, X_j)= \|\boldsymbol\mu_i-\boldsymbol\mu_j\|$,
\item $d_5(X_i, X_j)=
\frac{
|X_i|\,\mathrm{Mean}\left(
\left\{ \|\mathbf{x}_{u}-\boldsymbol\mu_i \|: \mathbf{x}_{u}\in X_i\right\}
\right)
+
|X_j|\,\mathrm{Mean}\left(
\left\{ \|\mathbf{x}_{v}-\boldsymbol\mu_j \|: \mathbf{x}_{v}\in X_j\right\}
\right)
}{
|X_i|+|X_j|
}$.
\end{itemize}
Bezdek and Pal in \cite{Dunn1974:index} considered also $d_6$ based
on the Hausdorff metric but this will be omitted here
as it turned out too slow to compute.

On the other hand, $D$ was chosen amongst:
\begin{itemize}
\item $D_1(X_i)=\mathrm{Max}\left(
\left\{ \|\mathbf{x}_{u}-\mathbf{x}_{v}\|: \mathbf{x}_{u},\mathbf{x}_{v}\in X_i\right\}
\right)$,
\item $D_2(X_i)=\mathrm{Mean}\left(
\left\{ \|\mathbf{x}_{u}-\mathbf{x}_{v}\|: \mathbf{x}_{u},\mathbf{x}_{v}\in X_i\right\}
\right)$,
\item $D_3(X_i)=\mathrm{Mean}\left(
\left\{ \|\mathbf{x}_{u}-\boldsymbol\mu_i\|: \mathbf{x}_{u}\in X_i\right\}
\right)$.
\end{itemize}

There are 15 different combinations of the possible numerators
and denominators in our study, hence 15 different CVIs,
which we will denote as \textit{GDunn\_dX\_DY}.
In particular, \textit{GDunn\_d1\_D1}
gives the original Dunn \cite{Dunn1974:index} index.

\subsection{CVIs based on near-neighbour graphs}

Let $\mathrm{NN}_M(i)=\{{j_1}, \dots, {j_M}\}$ denote
the set of the $i$-th point's $M$ nearest neighbours,
$0<\|\mathbf{x}_i-\mathbf{x}_{j_1}\|<\dots<\|\mathbf{x}_i-\mathbf{x}_{j_M}\|$
(assuming there are no tied distances, otherwise, some small random noise
can be added).

\paragraph{21--50) DuNN\_OWAs\_OWAc}
Note that the original Dunn index (denoted \textit{GDunn\_d1\_D1} above)
can be viewed as:
\begin{equation}
\mathrm{Dunn}(C)=\frac{
   \mathrm{Min}\left( \left\{ \|\mathbf{x}_i-\mathbf{x}_j\|: C(i)\neq C(j)\right\} \right)
}{
    \mathrm{Max}\left( \left\{ \|\mathbf{x}_i-\mathbf{x}_j\|: C(i) = C(j)\right\} \right)
}.
\end{equation}

Here we propose the following generalisation of the above --
a generalised Dunn-type index based on the notion of the
$M$-near-neighbour graph  and ordered weighted averaging
\cite{Yager1988:owa} operators -- convex combinations (weighted sums)
of ordered inputs. Namely:
\begin{equation}
\mathrm{DuNN}(C)=\frac{
    \mathrm{OWA}_s\left( \left\{ \|\mathbf{x}_i-\mathbf{x}_j\|: C(i)\neq C(j), i\in\mathrm{NN}_M(j)\text{ or }j\in \mathrm{NN}_M(i) \right\} \right)
}{
    \mathrm{OWA}_c\left( \left\{ \|\mathbf{x}_i-\mathbf{x}_j\|: C(i)= C(j), i\in\mathrm{NN}_M(j)\text{ or }j\in \mathrm{NN}_M(i) \right\}\right)
}.
\end{equation}
As a measure of cluster separation
we aggregate the ordered between-point distances
but only provided that they are part of the near-neighbour graph.
This will enable us to take the local point density into account
and detect well-separable clusters of even quite sophisticated  shapes.
In a similar manner, cluster compactness will be based on the nearest
neighbours as well.

Below we shall study pairs of $\mathrm{OWA}_s$ and $\mathrm{OWA}_c$
chosen amongst:
\begin{itemize}
\item $\mathrm{Min}$,
\item $\mathrm{Max}$,
\item $\mathrm{Mean}$,
\item
$\mathrm{SMin}_\delta(q_1,q_2,\dots,q_z)=
\sum_{i=1}^{z} w_{i,z} q_{(i)}$,  with
$w_{i, z}=\frac{\psi(i; z, \delta)}{\sum_{j=z-3\delta+1}^z \psi(j; z, \delta)}$
for $i> z-3\delta$ and 0 otherwise (``smooth minimum''),

\item
$\mathrm{SMax}_\delta(q_1,q_2,\dots,q_z)=
\sum_{i=1}^{z} w_{i,z} q_{(i)}$, with
$w_{i, z}=\frac{\psi(i; z, \delta)}{\sum_{j=1}^{3\delta} \psi(j; 1, \delta)}$
for $i\le 3\delta$ and 0 otherwise (``smooth maximum''),
\end{itemize}
where $q_{(1)} \ge q_{(2)} \ge \dots \ge q_{(z)}$
and $\psi(\cdot; \mu, \sigma)$ denotes the probability density function
of the normal distribution with expectation $\mu$ and standard deviation
$\sigma$, see also \cite{CenaGagolewski2020:genieowa}.

For instance, \textit{DuNN\_25\_SMin:5\_Max} denotes a generalised
Dunn index based on each point's 25 nearest neighbours.
It uses $\mathrm{SMin}_5$ as a separation measure
(computed over a subset of $25n$ distances restricted to the pair of points
belonging to different clusters)
and $\mathrm{Max}$ as a measure of compactness
(the remainder of the $25n$ distances comprised of point pairs
belonging to the same clusters).
Moreover, we will study indices like
\textit{DuNN\_25\_Mean\_Const}, where the denominator is fixed at 1.

In the sequel we will consider $M=5$ and $M=25$.
In order to keep the number of cases within reasonable limits,
we will restrict ourselves to 30 different CVIs of this type
(see Table~\ref{tab:res2} for a complete listing).

\paragraph{51,52) WCNN\_M}
The within-cluster near-neighbours (WCNN) index
is parametrised by  $M\ge 1$.
It aims to reflect how many
nearest neighbours of every point actually belong to the very same cluster:
\begin{equation}
\mathrm{WCNN}(C) =
\frac{
\left|C(i) = C(j):
j\in\mathrm{NN}_M(i)\right|
}{
nM
}.
\end{equation}
Ideally, $\mathrm{WCNN}(C) = 1$.
Hence, this is
a measure of how well the clusters are separated from each other.

Additionally, to prevent the formation of small clusters,
we will assume $\mathrm{WCNN}(C)=-\infty$ whenever
there is a cluster of cardinality $\le M$.
Similarly as above, we shall consider $M\in\{5, 25\}$.

%%%%%%%%%%%%%%%%%%%%%%%%%%%%%%%%%%%%%%%%%%%%%%%%%%%%%%%%%%%%%%%%%%%%%%%%%%%%%%%

\begin{table}[p!]
\caption{\label{tab:datasets} Benchmark datasets studied,
see \cite{clustering_benchmarks_v1}
and {https://github.com/gagolews/clustering\_benchmarks\_v1} for their
visual depictions; $l$ gives the number of reference partitions
and $k$s denote their possible cardinalities.}

\centering\footnotesize
\begin{tabularx}{1.0\linewidth}{rlrXrlrl}
\toprule
   & dataset                            & $k$s      & $l$  &    & dataset                            & $k$s & $l$   \\
\midrule
 1.& \textit{fcps/atom               }  & 2         &  1   & 32.& \textit{sipu/spiral             }  & 3    & 1     \\
 2.& \textit{fcps/chainlink          }  & 2         &  1   & 33.& \textit{sipu/unbalance          }  & 8    & 1     \\
 3.& \textit{fcps/engytime           }  & 2         &  2   & 34.& \textit{uci/ecoli               }  & 8    & 1     \\
 4.& \textit{fcps/hepta              }  & 7         &  1   & 35.& \textit{uci/ionosphere          }  & 2    & 1     \\
 5.& \textit{fcps/lsun               }  & 3         &  1   & 36.& \textit{uci/sonar               }  & 2    & 1     \\
 6.& \textit{fcps/target             }  & 2, 6      &  2   & 37.& \textit{uci/statlog             }  & 7    & 1     \\
 7.& \textit{fcps/tetra              }  & 4         &  1   & 38.& \textit{uci/wdbc                }  & 2    & 1     \\
 8.& \textit{fcps/twodiamonds        }  & 2         &  1   & 39.& \textit{uci/wine                }  & 3    & 1     \\
 9.& \textit{fcps/wingnut            }  & 2         &  1   & 40.& \textit{uci/yeast               }  & 10   & 1     \\
10.& \textit{graves/dense            }  & 2         &  1   & 41.& \textit{wut/circles             }  & 4    & 1     \\
11.& \textit{graves/fuzzyx           }  & 2, 4, 5   &  6   & 42.& \textit{wut/cross               }  & 4    & 1     \\
12.& \textit{graves/line             }  & 2         &  1   & 43.& \textit{wut/graph               }  & 10   & 1     \\
13.& \textit{graves/parabolic        }  & 2, 4      &  2   & 44.& \textit{wut/isolation           }  & 3    & 1     \\
14.& \textit{graves/ring             }  & 2         &  1   & 45.& \textit{wut/labirynth           }  & 6    & 1     \\
15.& \textit{graves/ring\_noisy      }  & 2         &  1   & 46.& \textit{wut/mk1                 }  & 3    & 1     \\
16.& \textit{graves/ring\_outliers   }  & 2, 5      &  2   & 47.& \textit{wut/mk2                 }  & 2    & 1     \\
17.& \textit{graves/zigzag           }  & 3, 5      &  2   & 48.& \textit{wut/mk3                 }  & 3    & 1     \\
18.& \textit{graves/zigzag\_noisy    }  & 3, 5      &  2   & 49.& \textit{wut/mk4                 }  & 3    & 1     \\
19.& \textit{graves/zigzag\_outliers }  & 3, 5      &  2   & 50.& \textit{wut/olympic             }  & 5    & 1     \\
20.& \textit{other/chameleon\_t4\_8k }  & 6         &  1   & 51.& \textit{wut/smile               }  & 4, 6 & 2     \\
21.& \textit{other/chameleon\_t5\_8k }  & 6         &  1   & 52.& \textit{wut/stripes             }  & 2    & 1     \\
22.& \textit{other/hdbscan           }  & 6         &  1   & 53.& \textit{wut/trajectories        }  & 4    & 1     \\
23.& \textit{other/iris              }  & 3         &  1   & 54.& \textit{wut/trapped\_lovers     }  & 3    & 1     \\
24.& \textit{other/iris5             }  & 3         &  1   & 55.& \textit{wut/twosplashes         }  & 2    & 1     \\
25.& \textit{other/square            }  & 2         &  1   & 56.& \textit{wut/windows             }  & 5    & 1     \\
26.& \textit{sipu/aggregation        }  & 7         &  1   & 57.& \textit{wut/x1                  }  & 3    & 1     \\
27.& \textit{sipu/compound           }  & 4, 5, 6   &  5   & 58.& \textit{wut/x2                  }  & 3    & 1     \\
28.& \textit{sipu/flame              }  & 2         &  2   & 59.& \textit{wut/x3                  }  & 4    & 1     \\
29.& \textit{sipu/jain               }  & 2         &  1   & 60.& \textit{wut/z1                  }  & 3    & 1     \\
30.& \textit{sipu/pathbased          }  & 3, 4      &  2   & 61.& \textit{wut/z2                  }  & 5    & 1     \\
31.& \textit{sipu/r15                }  & 8, 9, 15  &  3   & 62.& \textit{wut/z3                  }  & 4    & 1     \\
\bottomrule
\end{tabularx}
\end{table}

\section{Method}\label{sec:method}

\subsection{What is a valid cluster validity index?}

As we have proclaimed in the introduction,
our key assumption in this paper is that a meaningful
cluster validity measure $I$ should be high whenever
it is asked to assess the quality of one of the reference partitions,
and lower if it is applied on other clusterings.
In other words, useful CVIs should encourage the results
that agree with expert knowledge.

In order to be able to answer our main research question,
i.e., which cluster validity measures are valid,
we need the following components:
\begin{itemize}
\item benchmark data sets for evaluating the methods
(Section~\ref{sec:benchmarks}),
\item a procedure for finding the partition that maximises a given
CVI on each dataset (Section~\ref{sec:optim}),
\item a measure for quantifying the degree of agreement between what
a CVI thinks is a good partition vs~what experts have to say on this
matter (Section~\ref{sec:ari}).
\end{itemize}

\subsection{Benchmark Datasets}\label{sec:benchmarks}

We shall use an extensive battery of clustering benchmarks \cite{clustering_benchmarks_v1}\footnote{
Available at  \url{https://github.com/gagolews/clustering_benchmarks_v1}.
}, which not only combines data  that have already been used in a number of studies
\cite{fcps,uci,GravesPedrycz2010:kernelfuzzyclust,kmsix,xnn,chameleon},
but also features new test sets.

Most importantly, each benchmark dataset
comes with a set of $l\ge 1$ reference labels that were
assigned by experts. The case
$l>1$ reflects the situation where there might be multiple
valid/plausi\-ble/useful partitions (compare, e.g., \cite{multiple_clusterings});
we are dealing with an unsupervised learning problem after all.

The original benchmark battery consists
of 79 data instances, however 16 datasets are accompanied
by labels that yield $n(k-1) > 50{,}000$; they were omitted
for their computation would be too lengthy
(namely: \textit{mnist/digits},
\textit{mnist/fashion},
\textit{other/chameleon\_t7\_10k},
\textit{other/chameleon\_t8\_8k},
\textit{sipu/a1},
\textit{sipu/a2},
\textit{sipu/a3},
\textit{sipu/birch1},
\textit{sipu/birch2},
\textit{sipu/d31},
\textit{sipu/s1},
\textit{sipu/s2},
\textit{sipu/s3},
\textit{sipu/s4},
\textit{sipu/worms\_2},
\textit{sipu/worms\_64}).
Also \textit{uci/glass} has been removed as one of its 25-near-neighbour
graph's connected components was too small for the NN-based
methods to succeed. This leaves us with 62 datasets in total,
see Table~\ref{tab:datasets}.

Further, all columns of 0 variance were removed
and a tiny amount of noise (Gaussian with $\mu=0$ and $\sigma$ equal to
$10^{-6}$ of each column's sample standard deviation) was added so as
to assure the uniqueness of the clustering results.

\subsection{Finding optimal partitions (w.r.t.~a given CVI)}\label{sec:optim}

From now on we assume that the reader is familiar with the basics of the
language of mathematical programming, see, e.g., \cite{nocedal_wright,lee}
for a comprehensive overview.

For a predefined $\mathbf{X}$ and $k$,
let us fix a cluster validity measure $I:\mathcal{C}_k\to\mathbb{R}$. Without loss in generality,
we assume that the higher the $I$, the more useful the partition.
This is because we can always take $I:=-I$, as we have done
with the Ball--Hall and Davies--Bouldin indices above.

For a given $k$-partition $C$, let
$\mathrm{NEIGHBOURS}(C)$ denote
the set of all surjections like $C': [1:n]\stackrel{\text{onto}}{\to}[1:k]$
with $C'(i)\neq C(i)$ for some $i$ and $C'(j)=C(j)$ for all $j\neq i$.
In other words, it is the set of all $k$-partitions that can be obtained from
$C$ by relocating a single point to some other cluster.

\medskip
We are interested in finding a partition which is a solution
to the optimisation problem:
\begin{equation}\label{eq:optim}%\tag{$\star$}
\maximise_{C\in\mathcal{C}_k} I(C),
\end{equation}
i.e., $C^*\in\mathcal{C}_k$ such that
$I(C^*)\ge I(C)$ for all $C\in\mathcal{C}_k$.

\begin{remark}
The solution to \eqref{eq:optim} is not unique;
clusterings are defined up to a permutation of the cluster numbers (IDs).
For example, a 2-partition of a 4-ary set encoded
like $(C(1), C(2), C(3), C(4))=(1, 1, 2, 1)$
is semantically equivalent to $(2, 2, 1, 2)$.
Moreover, it might happen that a dataset exhibits a number of equally good
splits. This is exactly the case when we apply \textit{WCNN\_$M$}
on datasets whose $M$-near-neighbour graphs are disconnected
and the number of connected components is greater than $k$.
For instance, assuming $Y_1,Y_2,Y_3$ are disconnected, in this setting
the 2-partition $\{Y_1\cup Y_2, Y_3\}$ is as good as $\{Y_1, Y_2\cup Y_3\}$.
\end{remark}

\bigskip
In general, the combinatorial optimisation problem \eqref{eq:optim}
is extremely difficult to solve in practice.
Enumerating all the possible solutions is virtually impossible
as the number of possible partitions is equal to the Stirling number
of the second kind,
$S(n,k)=\frac{1}{k!} \sum_{i=0}^k (-1)^i {k \choose i} (k-i)^n$
which is $O(k^n)$, and note that in our case $2\le k\ll n$.

In this paper, however, we shall make \textit{reasonable efforts}
towards finding the maximum of the objective \eqref{eq:optim}.
In essence, for each dataset we will generate dozens of
``interesting'' partitions (using existing state-of-the art clustering algorithms and evolutionary-based heuristics, see Section~\ref{sec:candidates})
each of which we shall then try to improve
with an expansive variant of the steepest ascent hill climbing
(with tabu \cite{tabu} search-like memoisation)
that itself guarantees to land in a local maximum of the objective \eqref{eq:optim}.

\bigskip
The maximum of the objective \eqref{eq:optim} will be sought by means
of the following variant of the hill climbing scheme.

\begin{algorithm}\label{algo:tabu}
With $\{C_1,\dots,C_m\}$ let us denote the set of
initial candidate solutions (see Section \ref{sec:candidates}
for more details) ordered in such a way that
$I(C_1)\ge\dots\ge I(C_m)$.
For brevity of notation, we assume that $I(\emptyset)=-\infty$.

\begin{enumerate}
\item[In:] $I:$, $C_1,C_2,\dots,C_m$, $P\in\mathbb{N}$;
\item[1.] $\mathcal{T} = \emptyset$; \hfill\textit{(a ``tabu'' list)}
\item[2.] $C^* = C_1$; \hfill\textit{(best solution so far)}
\item[3.] \textbf{for} $C = C_1,C_2,\dots,C_m$ \textbf{do}: \hfill\textit{($I(C_1)\ge\dots\ge I(C_m)$)}
    \begin{enumerate}
    \item[3.1.] $p=1$;
    \item[3.2.] $C^+ = \emptyset$;
    \item[3.3.] \textbf{for each} $C' \in \mathrm{NEIGHBOURS}(C)$ \textbf{do}:
        \begin{enumerate}
        \item[3.3.1.] \textbf{if} $C'\not\in \mathcal{T}$ and $I(C') > I(C^+)$, \textbf{then} $C^+ = C'$;
        \end{enumerate}
    \item[3.4.] \textbf{if} $C^+ = \emptyset$ \textbf{then} continue to step 3; \hfill\textit{(cannot improve further)}
    \item[3.5.] $\mathcal{T} = \mathcal{T} \cup \{ C^+ \}$; \hfill\textit{(never visit $C^+$ again)}
    \item[3.6.] $C = C^+$;
    \item[3.7.] \textbf{if} $I(C) > I(C^*)$, \textbf{then} $C^* = C$, \textbf{else} $p = p+1$;
    \item[3.8.] \textbf{if} $p \le P$, \textbf{then} go to step 3.2; \hfill\textit{(try to improve current $C^+$ next)}
    \end{enumerate}
\item[4.] \textbf{return} $C^*$;
\end{enumerate}
\end{algorithm}

It is easily seen that the algorithm guarantees that the solution returned
cannot be further improved
by relocating an individual point to a different cluster.
Hence, the return value
is definitely a local maximum,
however there is of course no guarantee that the identified
optimum is global.
As we argue below, though, it will turn out sufficient for our purposes.

We shall set the upper bound for the number of iterations without improvement,
$P$, to $250$ (we have rarely seen any improvements beyond $100$, though).
This allows for the procedure to explore
the area around the candidate solutions quite broadly.
Note that the $\mathcal{T}$ set, which guarantees that no partition
is considered twice, is shared across all the iterations
so that the visited subspace is even broader.
Hence, the search is more comprehensive than if we had
restarted the whole procedure independently for each $C_1,\dots,C_m$ and
then chose the best amongst the identified local maxima.

\begin{remark}
Note that the number of points in $\mathrm{NEIGHBOURS}(C)$
is $O(n(k-1))$.
Overall, the procedure for certain CVIs can be sped up by computing
$I(C^+)$ incrementally based on $I(C')$ and the knowledge of which point is
being relocated to which cluster.
For instance, the Silhouette index only requires an $O(nk)$
update instead of a full recompute worth of $O(n^2)$ time.
The CVIs we have considered gave the time complexity of steps 3.1--3.8
most often lying between $O(n^2 k^2)$ and $O(n^3 k^2)$
(with constants $d$ and $M$ having some obvious influence as well).
The typical size of $\mathcal{T}$ at the end of the algorithm's run
(i.e., the number of executions of step 3.5)
when started from $m=5$ random points was 1000--2000.
\end{remark}

\subsection{Measuring similarity to reference partitions}\label{sec:ari}

For a given benchmark dataset $\mathbf{X}$,
let $C_1^\$, C_2^\$, \dots, C_l^\$$ be the reference partitions
and $k_1,\dots,k_l$ be their respective cardinalities.

Note that any clustering method $\mathfrak{c}$
(for example, the maximiser  of the Caliński--Harabasz index
or the Ward linkage) can be thought of as a function
that takes $\mathbf{X}$ and $k_i$ on input and yields a $k_i$-partition
of $\mathbf{X}$ on output, i.e.,
$\mathfrak{c}(\mathbf{X}, k_i)\in\mathcal{C}_{k_i}$.

We will use the adjusted Rand index (ARI)
\cite{HubertArabie1985:partitionscomp,psi}
to measure the similarity between $\mathfrak{c}(\mathbf{X}, k_i)$
and $C_i^\$$.
Recall that two equivalent partitions yield ARI equal to 1.
Moreover, two ``independent'' (see \cite{ari_adjustment}
for discussion) clusterings have the expected ARI of 0.
Negative ARIs will be replaced with 0 for better interpretability of the results.

\medskip
In order to quantify the quality of the method $\mathfrak{c}$  on $\mathbf{X}$,
we will evaluate its outputs against all the available reference labellings
and choose the highest ARI in result:
\begin{equation}
Q_\mathbf{X}(\mathfrak{c}) = \max\left\{
\mathrm{ARI}\left(\mathfrak{c}(\mathbf{X}, k_1), C_1^\$\right),
\dots,
\mathrm{ARI}\left(\mathfrak{c}(\mathbf{X}, k_l), C_l^\$\right)
\right\}.
\end{equation}
This is to account for the fact that
there might be many equally valid partitions
and the (unsupervised) method $\mathfrak{c}$ should be rewarded
if it identifies one of them (does not matter which one).

It is worth noting that only 13 datasets have $l>1$, 11 of which
come with reference labellings that do not have identical cardinalities.
Also, some reference partitions include noise points --
these were excluded during the computations of the ARIs
(after the output of $\mathfrak{c}$ was determined,
as none of the clustering methods studied features a noise point detector).
Overall, this validation methodology conforms with \cite{clustering_benchmarks_v1}.

\subsection{Candidate solutions}\label{sec:candidates}

To generate the list of candidate (initial) solutions
used in Algorithm~\ref{algo:tabu},
we will apply many different clustering algorithms
on each dataset, including:
\begin{itemize}
\item the most popular hierarchical clustering methods
(single, average, Ward, centroid, complete linkage),
\item Genie \cite{GagolewskiETAL2016:genie} (with different thresholds),
\item information-theoretic algorithms (ITM \cite{itm} as well as
IcA and GIc \cite{Cena2018:phd}),
\item other methods in the well-established
\texttt{sklearn} \cite{sklearn} package for Python:
$k$-means, Gaussian mixtures, spectral clustering with different kernels,
Birch (with a range of parameter values).
\end{itemize}
This gives 87 different combinations of algorithms and their setups.
In Section \ref{sec:experiment} we provide some technical details
about their implementations. Note that 12 of them will constitute
the baseline in our empirical study below.

Also, we shall utilise the following heuristic solvers:
\begin{itemize}
\item particle swarm optimisation (via R package \texttt{pso} \cite{rpso}),
\item ``global'' optimisation by differential evolution \cite{deoptim}
(\texttt{DEoptim} \cite{rdeoptim} in R).
\end{itemize}
They  pinpoint local maxima based on 3--5 restarts from
different initial candidate solutions.
Both of them search over the continuous space $\mathbb{R}^{(Vk)\times d}$
in such a way that the clusters are represented by means of $Vk$
vantage points. $V$ vantage points represent one cluster
(empirically, we have determined $V=5$ be a good compromise between
quality and speed).
In every iteration, each point is assigned to its closest vantage point, and,
as a consequence, to a cluster which is represented by this vantage point.
This approach allows to determine clusters of more sophisticated shapes
than when simply $V=1$ is utilised (as with $V>1$ we consider different unions of cells in a Voronoi diagram).

Additionally, to broaden the search space even further,
we will pick 5 partitions completely at random,
i.e., each $C\in\mathcal{C}_k$ being
such that $C(1),\dots,\allowbreak C(n)$
being independent random variables
from the discrete uniform distribution on $[1:k]$.
Nevertheless, starting from a random partition never turned out better than
an assisted initialisation based on one of the aforementioned
candidate solutions.

Most importantly, as each dataset comes with a set of reference labels given
by experts, these shall be considered as well.

\bigskip
Overall, for each dataset, we have obtained $m\simeq 100$
different clusterings%
\footnote{All results are available at \url{https://github.com/gagolews/clustering_results_v1}.},
however, quite often there were duplicated entries,
hence the effective $m$ was in the range 30--50.

\bigskip
An ideal index, if it existed, would be 100\% concordant
with expert labels. That is, it would be impossible for the hill climbing
method to improve them any further. Note that when we maximise $I$,
the reference partitions are always amongst the initial candidate
solutions which are fed to Algorithm~\ref{algo:tabu}.
Therefore, our procedure guarantees that all the good combinations
of indices and datasets must be identified.
If the hill climbing method converges to a different solution,
it means that $I$ promotes some points that are less compatible with
experts' opinion.

Let us stress that neither $I$ itself, nor
the procedure for maximising it, is ``aware'' of
the existence of any external labels: only $\mathbf{X}$ and $k$
are input to $\mathfrak{c}$, not $C_i^\$$; this is still an unsupervised
learning method.
Algorithm~\ref{algo:tabu} converges where it converges;
the starting points are plentiful
and there is a great variety of them.
Reference partitions are only used at the final evaluation stage.

\section{Experiments}\label{sec:experiment}

\subsection{Implementation}

Experiments, data analysis, and visualisation tasks were performed using
Python 3.8.6 (PyPI packages:
\texttt{numpy} 1.19.0,
\texttt{scipy} 1.5.1,
\texttt{pandas} 1.0.3,
\texttt{matplotlib} 3.3.3,
\texttt{seaborn} 0.11.1)
% below:
% sklearn 0.23.1
% fastcluster 1.1.26
% genieclust 0.9.4
% ITM 178fd43
and R 4.0.3 (CRAN packages: \texttt{DEoptim} 2.2-5 \cite{rdeoptim},
\texttt{pso} 1.0.3 \cite{rpso}).

The correctness of our C++ implementations%
\footnote{Available at \url{https://github.com/gagolews/optim_cvi}.}
of the cluster validity
indices was verified against R packages
\texttt{clusterCrit} 1.2.8 and \texttt{clusterSim} 0.49-2 (wherever applicable).
Our library turned out significantly faster than the two reference ones.
Moreover, we allowed for the computing of the indices incrementally
(as mentioned above), which was particularly beneficial
in terms of the run-time of Algorithm \ref{algo:tabu}.

Overall, the computations took ca.~3 months of computing time
with the use of 2 computer clusters
(within the allocated resource limits we have been granted
by ICM UW/PL-Grid and the School of IT at Deakin University).

\subsection{Which index best agrees with expert knowledge?}

Let us proceed with the evaluation of the agreement between the
cluster validity indices and expert knowledge.

\subsubsection{GDunn\_dX\_DY}
We first focus on the 15 generalised Dunn indices \cite{BezdekPal1998:gdunn}.
To recall, \textit{GDunn} indices are
defined as the ratio of cluster separation ($d$)
and compactness ($D$).

\begin{figure}[h!]
\centering
\includegraphics[width=1.0\linewidth]{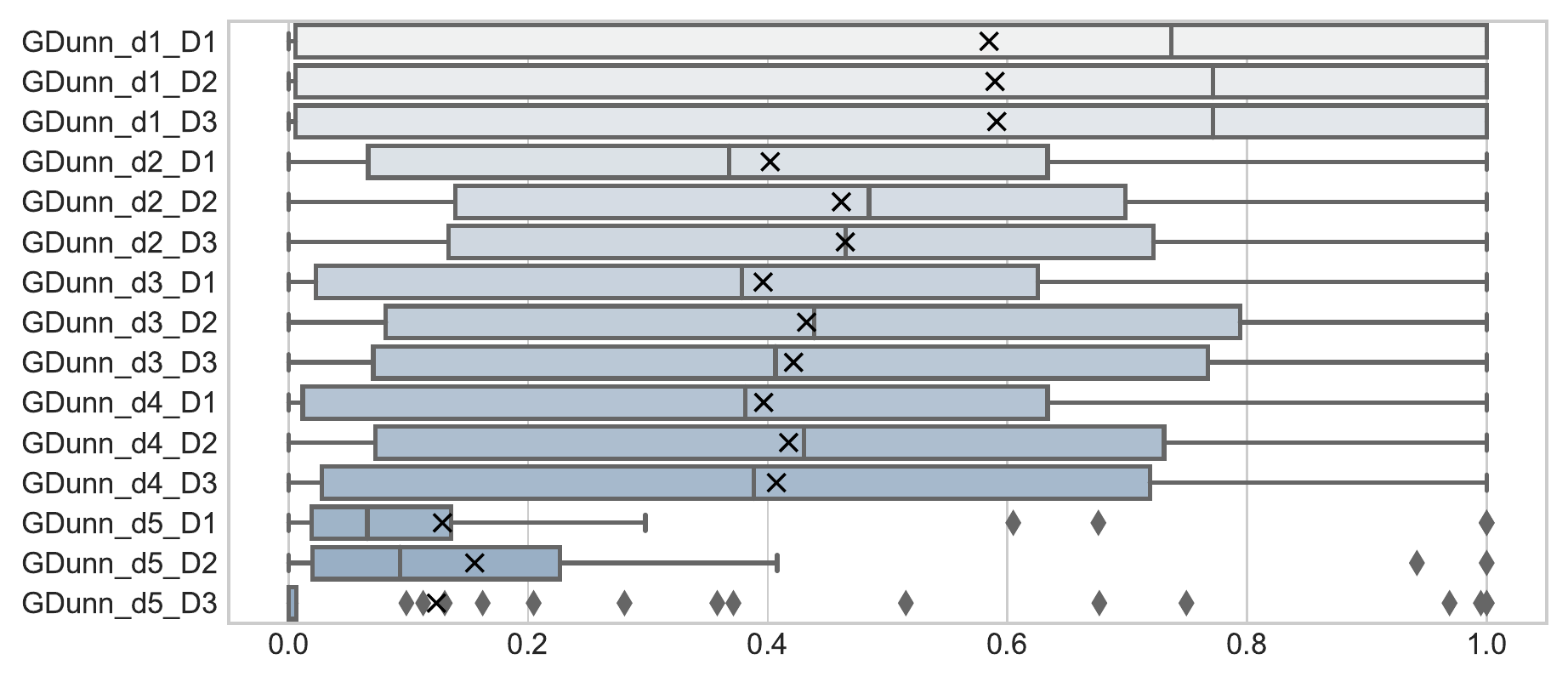}
\caption{\label{fig:boxplot-ar-GDunn} ARI: Generalised Dunn indices.
We see that the choice of the cluster compactness measure $D$ is rather negligible. On the other hand, separation measure $d_1=\mathrm{Min}$
performs best whilst $d_5$ (averaged distance to cluster centres) is subpar.}
\end{figure}

Figure~\ref{fig:boxplot-ar-GDunn} gives the box-and-whisker plots
for the Adjusted Rand indices across the benchmark datasets studied.
We clearly see that $d_1$, i.e., the pairwise minimum distance
(used in the original Dunn index) outperforms the other measures.
In this scenario, the Wilcoxon signed-rank test does not
find the choice of $D$ significant ($\alpha=0.05$).

Also, the measures based on $d_5$ are significantly worse than
all other ones. Perhaps it would be better
if $d_5$ was defined as the average \textit{squared}
point-centroid distance, not just average raw distance;
recall that a centroid is the point that minimises exactly
the square of the Euclidean metric.

\subsubsection{DuNN}
\begin{figure}[h!]
\centering
\includegraphics[width=1.0\linewidth]{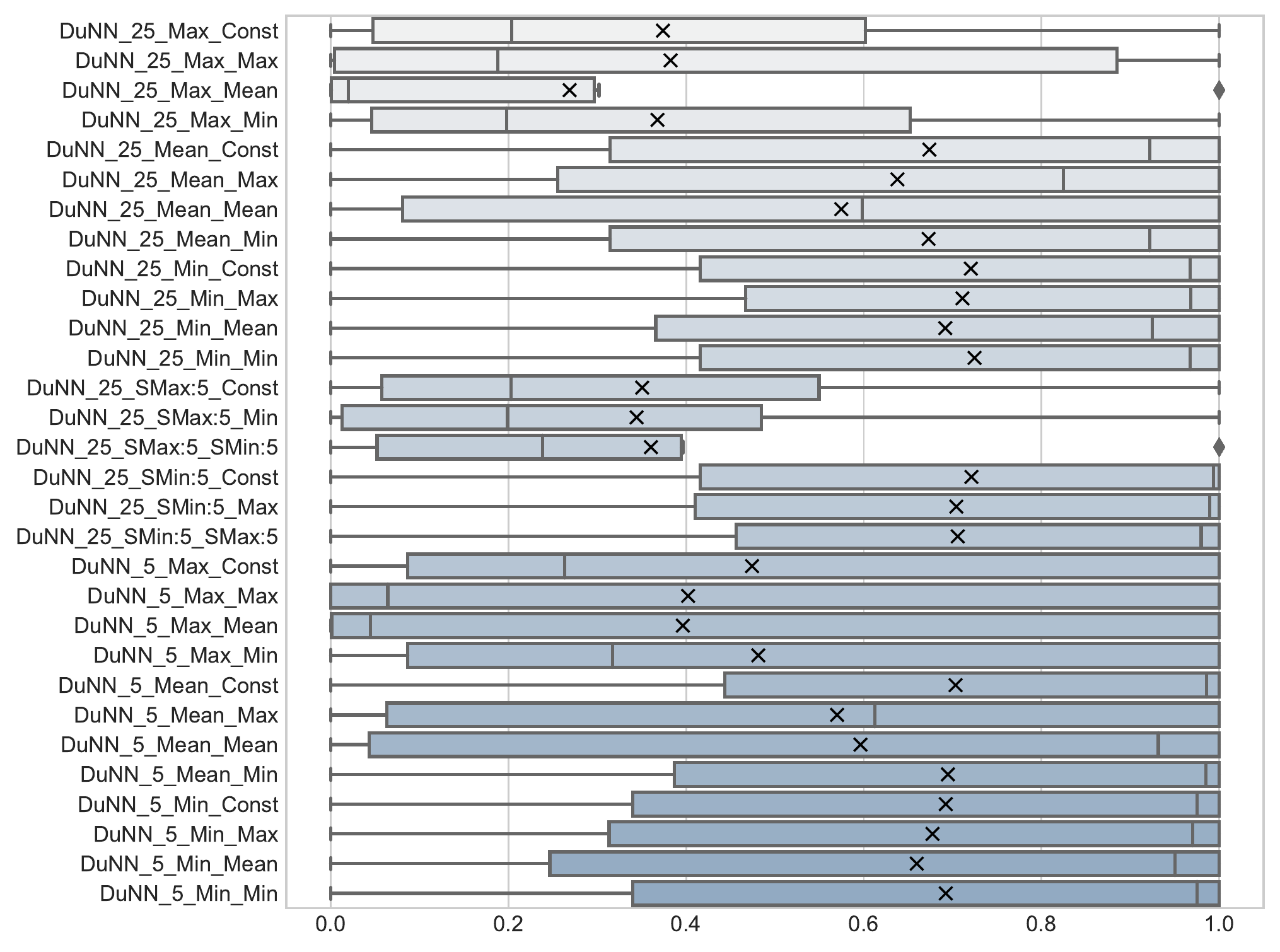}
\caption{\label{fig:boxplot-ar-DuNN} ARI: Near-neighbour-based DuNN indices.
Disabling the use of a compactness measure whatsoever  (\textit{*\_Const})
might be preferred. Also, it is better to
have the closest pairs of points from different clusters as far away
from each other as possible.}
\end{figure}

Figure~\ref{fig:boxplot-ar-DuNN} shows the empirical distribution of ARIs
in the case of the near-neighbour versions of the Dunn index.

For a fixed separation measure $\mathrm{OWA}_s$,
$\mathrm{OWA}_c$ equal to $\mathrm{Min}$ and $\mathrm{Const}$ is never
significantly worse (one-sided Wilcoxon test, $\alpha=0.05$)
than $\mathrm{Max}$ and $\mathrm{Mean}$.
Moreover, there is no significant difference between
$\mathrm{Min}$ and $\mathrm{Const}$, therefore,
applying Ockham's razor,
we conclude that the cluster compactness could be omitted whatsoever
(at least as far as our selection of aggregation functions is concerned).

Setting $\mathrm{OWA}_c$ at $\mathrm{Const}$,
interestingly,
\textit{DuNN\_25\_SMin:5\_Const}
significantly outperforms all the variants except
\textit{DuNN\_5\_Mean\_Const}.

Moreover, \textit{DuNN\_25\_Min\_Const}
is better than \textit{DuNN\_5\_Min\_Const}.
Also note that the behaviour of \textit{Max} or its smoothened
version is particularly poor.

\subsubsection{Other methods}
As a base line, the above and remaining CVIs will be compared against
the outputs of 12 clustering algorithms:
\begin{enumerate}
\item[1--5)] \textit{Average, Centroid, Complete, Ward, Single} -- classical agglomerative
hierarchical clustering algorithms;
\item[6--9)] \textit{Genie\_G0.1, Genie\_G0.3, Genie\_G0.5, Genie\_G0.7} -- the robust
hierarchical clustering algorithm Genie that we have proposed in \cite{GagolewskiETAL2016:genie},
with different thresholds for the Gini index of the inequity in cluster sizes;
\item[10)] \textit{ITM} -- greedy divisive minimiser of an information theoretic criterion
over minimum spanning trees \cite{itm};
\item[11)] \textit{GaussMix} -- expectation-maximisation (EM) for Gaussian mixtures with 100 restarts and each cluster having its own covariance matrix;
\item[12)] \textit{KMeans} -- Lloyd-like $k$-means algorithm with 10 restarts
(note that this is a heuristic to optimise the Caliński--Harabasz
index/within-cluster sum of squares).
\end{enumerate}
Their implementations are included in Python packages (available via PyPI;
see their respective API documentation for more details on algorithms
and default values of their parameters in place)
\texttt{fastcluster} 1.1.26 (\textit{Average, Centroid, Complete, Ward}; \cite{Mullner2013:fastcluster}),
\texttt{genieclust} 0.9.4 (\textit{Genie\_G0.x, Single}; \cite{genieclust}),
\texttt{sklearn} 0.23.1 (\textit{GaussMix, KMeans}; \cite{sklearn}).
Moreover, the implementation of \textit{ITM} \cite{itm}
is available from GitHub\footnote{
See \url{https://github.com/amueller/information-theoretic-mst};
git commit 178fd43.}.

\begin{figure}[p!]

\vspace*{-3cm}  % sorry about that :)
\centering
\includegraphics[width=1.0\linewidth]{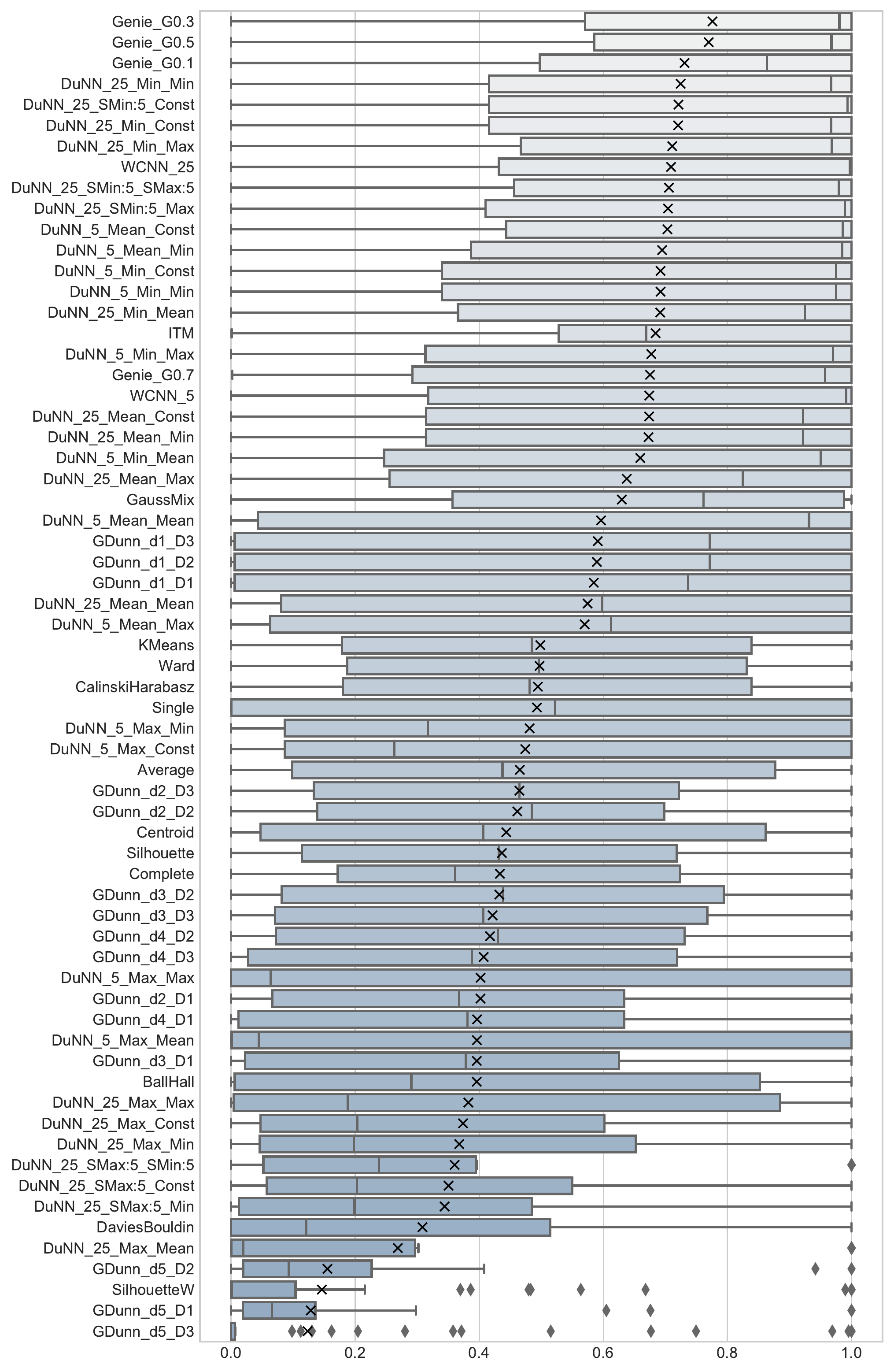}
\caption{\label{fig:boxplot-ar-all} ARI: All methods (ordered by the average ARI). The Genie algorithm outperforms other clustering approaches.
Near-neighbour-based cluster validity measures reflect
expert knowledge quite well.}
\end{figure}

Tables \ref{tab:res1} and \ref{tab:res2} give  the basic summary statistics
on the empirical distribution of the ARIs across the 62 benchmark datasets
and all the methods studied. Moreover, Figure~\ref{fig:boxplot-ar-all}
displays the boxplots.

We observe what follows:
\begin{itemize}
\item Our Genie algorithm \cite{genieclust} outperforms
other methods. Note that it is significantly faster than
most other algorithms as it is based on the minimum spanning tree
of the pairwise distance graph.
\item The lesser-known ITM method \cite{itm} performs relatively well.
\item The near-neighbour-based \textit{DuNN} indices that we have
proposed in this paper are much better than \textit{GDunn}.
\item The difference between \textit{DuNN\_25\_SMin:5\_Const}
and \textit{WCNN\_25} is insignificant.
\item Overall, the algorithms based on the near-neighbourhood (minimum spanning
trees can be considered a variant thereof, in some sense),
seem to be much more valid for clustering tasks.
\item Next in line are Gaussian mixtures that detect clusters of
specific (spherical) shapes.
\item Of course, $k$-means give similar results to the Caliński--Harabasz
optimiser as the former is a heuristic to optimise the latter as the objective.
Also \textit{Ward} is a greedy agglomerative maximiser of the same objective.
\item Most other cluster validity measures seem to promote the
clusterings that are not concordant to expert knowledge
which calls their relevance into question.
\item \textit{SilhouetteW} and \textit{BallHall} -- the
weighted-by-cluster-cardinality versions
of \textit{Silhouette} and \textit{CalińskiHarabasz}, respectively,
perform worse than their unadjusted counterparts.
\item The poor performance of some methods may be partially explained by
the inequality of the cluster sizes they output -- some of them
are prone to generating few very large clusters and a number of very small
ones (perhaps even being singleton objects).
This includes \textit{Single} (median Gini index of the cluster sizes=0.85),
\textit{DaviesBouldin} (0.95),
and \textit{SilhouetteW} (0.98).
On the other hand,
\textit{Dunn\_25\_Max\_Mean} (median Gini index of the cluster sizes=0.09),
\textit{Dunn\_5\_Max\_Mean} (0.09),
\textit{CalińskiHarabasz} (0.13),
\textit{KMeans} (0.15),
\textit{ITM} (0.17),
and \textit{Genie\_G0.1} (0.17)
produced the least \textit{skewed} partition sizes.
However, let us note that this is not necessarily an accurate predictor
of the clustering quality (see also \cite{genieclust} for discussion).
\item Note that some datasets are inherently hard to cluster
(the outputs of no algorithm matches the reference partition well);
these include
\textit{uci/sonar}  (max ARI=0.036),
\textit{uci/yeast}  (max ARI=0.181), and
\textit{uci/iono\-sphere}  (max ARI=0.401).

\end{itemize}

\subsection{Clustering of clustering algorithms}

Let us perform an interesting exercise
where we determine a grouping of the clustering methods by means of the
overall similarity of the results they generate on all the benchmark datasets.
This way, we will know which methods (and CVIs) are
``semantically'' similar to each other.

We have computed the AR indices between all pairs of label vectors
generated by all the methods (this time, reference/expert labels
were not used). We used the mean, the median, or
the 3rd quartile of $(1.0-\textrm{ARI})$
to obtain a single number that summarises the ``distance''
between the algorithms.

\begin{figure}[p!]
\centering

\vspace*{-2cm}
\hspace*{-0.15\linewidth}
\begin{minipage}{1.3\linewidth}
\includegraphics[width=0.31\linewidth]{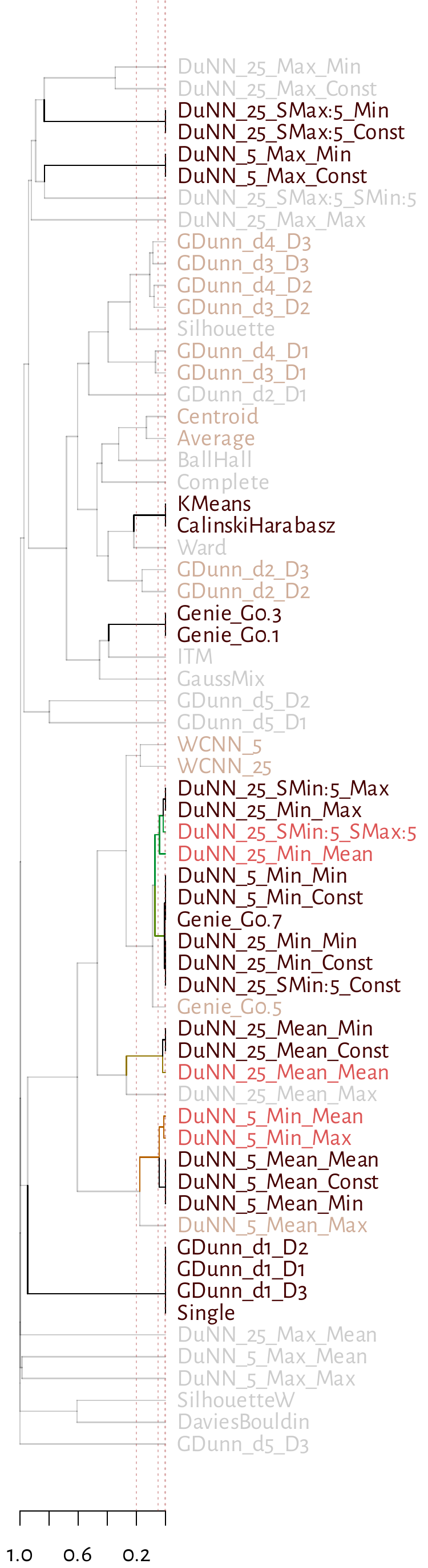}
\includegraphics[width=0.31\linewidth]{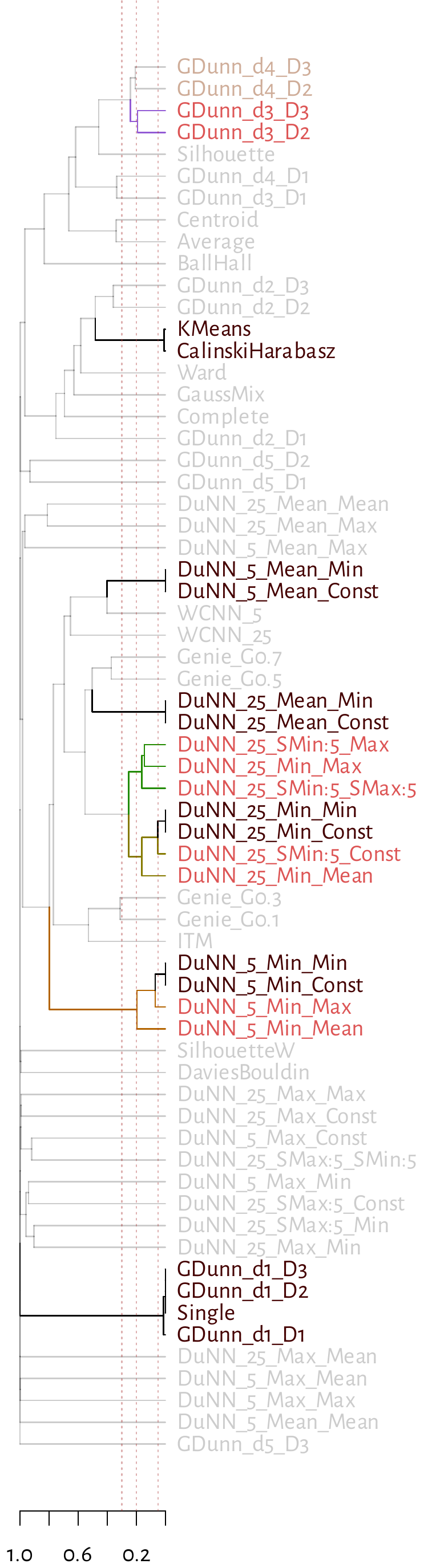}
\includegraphics[width=0.31\linewidth]{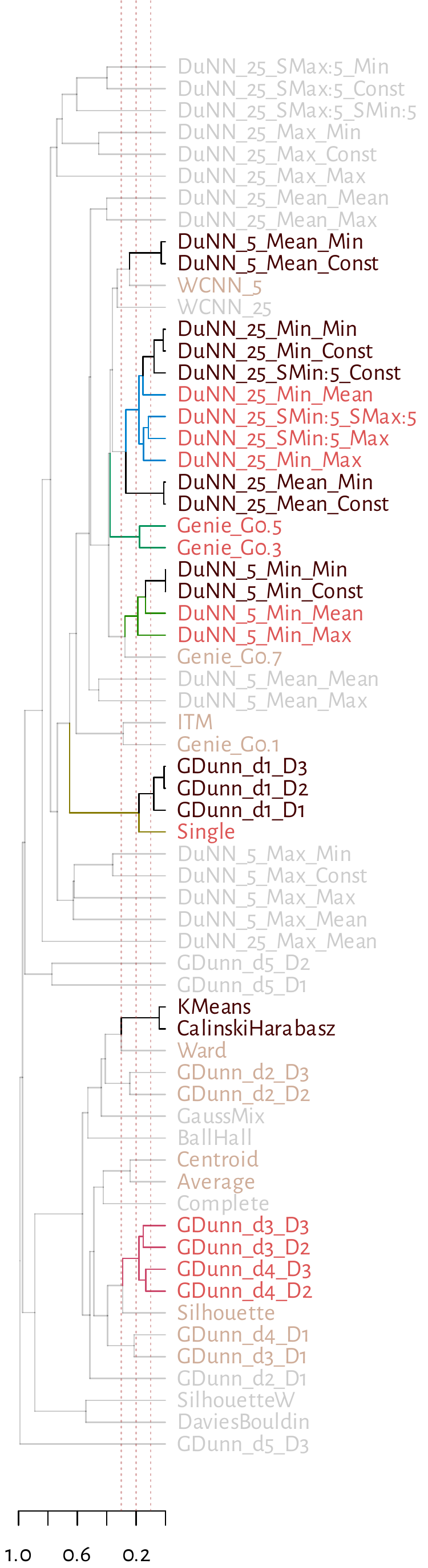}
\end{minipage}
\caption{\label{fig:pairs-complete}
Complete linkage with Median(1.0-ARI), Q3(1.0-ARI), Mean(1.0-ARI), respectively.
Only clusters that are formed at low dissimilarity levels (branches
with connectors in the right parts of each figure) can be considered
meaningful.}
\end{figure}

We have applied the agglomerative hierarchical clustering algorithm
with complete linkage so that the resulting dendrograms,
which are depicted in Figure~\ref{fig:pairs-complete},
are more interpretable
(this will give us the maximal aggregated dissimilarities
between all methods in a cluster).
Note that we will be only interested in groups of algorithms
that have small pairwise distances.

\bigskip
The majority vote of the results obtained by means
of all the three dissimilarity measures gives the following ``consensus''
clusters, where the algorithms have quite high overall degree of similarity:

\begin{itemize}
% ## [1] "CalińskiHarabasz, KMeans"
% ##  [1] "CalińskiHarabasz, KMeans"
% ## [1] "CalińskiHarabasz, KMeans"
\item \textit{CalińskiHarabasz, KMeans},

% ## [2] "DuNN_25_Mean_Const, DuNN_25_Mean_Min"
% ## [2] "DuNN_25_Mean_Const, DuNN_25_Mean_Min"
% ##  [3] "DuNN_25_Mean_Const, DuNN_25_Mean_Min"
\item \textit{DuNN\_25\_Mean\_Const, DuNN\_25\_Mean\_Min},

% ##  [8] "DuNN_5_Min_Const, DuNN_5_Min_Min, Genie_G0.7"
% ##  [5] "DuNN_5_Min_Const, DuNN_5_Min_Min"
% ##  [5] "DuNN_5_Min_Const, DuNN_5_Min_Min"
\item \textit{DuNN\_5\_Min\_Const, DuNN\_5\_Min\_Min},

% ## [3] "DuNN_25_Min_Const, DuNN_25_Min_Min, DuNN_25_SMin:5_Const"
% ##  [4] "DuNN_25_Min_Const, DuNN_25_Min_Min, DuNN_25_SMin:5_Const"
% ## [3] "DuNN_25_Min_Const, DuNN_25_Min_Min"
\item \textit{DuNN\_25\_Min\_Const, DuNN\_25\_Min\_Min, DuNN\_25\_SMin:5\_Const},

% ##  [7] "DuNN_5_Mean_Const, DuNN_5_Mean_Mean, DuNN_5_Mean_Min"
% ##  [4] "DuNN_5_Mean_Const, DuNN_5_Mean_Min"
% ##  [4] "DuNN_5_Mean_Const, DuNN_5_Mean_Min"
\item \textit{DuNN\_5\_Mean\_Const, DuNN\_5\_Mean\_Min},

% ## [6] "GDunn_d1_D1, GDunn_d1_D2, GDunn_d1_D3"
% ## [6] "GDunn_d1_D1, GDunn_d1_D2, GDunn_d1_D3, Single"
% ## [10] "GDunn_d1_D1, GDunn_d1_D2, GDunn_d1_D3, Single"
\item \textit{GDunn\_d1\_D1, GDunn\_d1\_D2, GDunn\_d1\_D3, Single}.

\end{itemize}

\bigskip
Let us note that:
\begin{itemize}
\item The above indicates again that in our task, the denominator
(compactness measure) in all the generalisations of the Dunn index
has no significant impact on the results.
However, when CVIs are applied for the purpose of selecting
the optimal number of clusters, the conclusions could be much different.

\item \textit{GDunn} and \textit{DuNN} are not so similar, despite
they both generalise the same index, \textit{Dunn}.
On the other hand, the latter (to recall, it is based on $d_1=\mathrm{Min}$)
is similar to \textit{Single} linkage (which is determined
through a greedy (agglomerative) consumption of the
nearest pairs of points; it can be computed based on a minimum
spanning tree).

\item \textit{Ward} is quite similar to \textit{CalińskiHarabasz},
and \textit{KMeans}, which was to be expected as they tend to optimise
the same objective function.

\end{itemize}

%%%%%%%%%%%%%%%%%%%%%%%%%%%%%%%%%%%%%%%%%%%%%%%%%%%%%%%%%%%%%%%%%%%%%%%%%%%%%%%

\section{Conclusion and Future Work}\label{sec:conclusion}

We have studied whether cluster validity indices
really promote partitions that reflect what experts judge
as meaningful.
While some measures could still be considered relevant in the task of selecting
the right number of clusters,
it is better not to treat them as objective functions
for identifying good partitions. This is particularly the case
with the Davies--Bouldin, Silhouette, Ball--Hall,
and the no-near-neighbour-based versions of the generalised Dunn
index.

In the future, we will verify the usability of our new
near-neighbour-based generalisations of the Dunn index
in the problem of choosing the meaningful number of clusters.
Their advantage is that they take into account the locality
of the input points as well as the relative density of the points'
distribution. Certainly, more combinations of OWA operators
should be studied.

%%%%%%%%%%%%%%%%%%%%%%%%%%%%%%%%%%%%%%%%%%%%%%%%%%%%%%%%%%%%%%%%%%%%%%%%%%%%%%%

\section*{Acknowledgements}

\noindent
This  research was supported by the Australian Research Council Discovery
Project ARC DP210100227
as well as the PL-Grid Infrastructure.

\section*{Conflict of interest}

\noindent
The authors declare that they have no conflict of interest to disclose.

%%%%%%%%%%%%%%%%%%%%%%%%%%%%%%%%%%%%%%%%%%%%%%%%%%%%%%%%%%%%%%%%%%%%%%%%%%%%%%%

\begin{table}[p!]
\caption{\label{tab:res1} ARI: Basic summary statistics; part I.
}

\smallskip\centering
\begin{tabular}{lrrrrr}
\toprule
Method                      {} &   Mean &   St.Dev. &    Q1 &   Median &   Q3 \\
\midrule
\textit{Average                  }&   0.47 &  0.38 &  0.10 &  0.44 &  0.88 \\
\textit{Centroid                 }&   0.44 &  0.39 &  0.05 &  0.41 &  0.86 \\
\textit{Complete                 }&   0.43 &  0.33 &  0.17 &  0.36 &  0.72 \\
\textit{GaussMix                 }&   0.63 &  0.38 &  0.36 &  0.76 &  0.99 \\
\textit{Genie\_G0.1              }&   0.73 &  0.32 &  0.50 &  0.86 &  1.00 \\
\textit{Genie\_G0.3              }&   0.78 &  0.29 &  0.57 &  0.98 &  1.00 \\
\textit{Genie\_G0.5              }&   0.77 &  0.32 &  0.59 &  0.97 &  1.00 \\
\textit{Genie\_G0.7              }&   0.68 &  0.38 &  0.29 &  0.96 &  1.00 \\
\textit{ITM                      }&   0.68 &  0.28 &  0.53 &  0.67 &  1.00 \\
\textit{KMeans                   }&   0.50 &  0.35 &  0.18 &  0.48 &  0.84 \\
\textit{Single                   }&   0.49 &  0.47 &  0.00 &  0.52 &  1.00 \\
\textit{Ward                     }&   0.50 &  0.35 &  0.19 &  0.50 &  0.83 \\
\midrule
\textit{BallHall                 }&   0.40 &  0.40 &  0.01 &  0.29 &  0.85 \\
\textit{CalińskiHarabasz         }&   0.49 &  0.35 &  0.18 &  0.48 &  0.84 \\
\textit{DaviesBouldin            }&   0.31 &  0.37 &  0.00 &  0.12 &  0.51 \\
\textit{Silhouette               }&   0.44 &  0.37 &  0.11 &  0.43 &  0.72 \\
\textit{SilhouetteW              }&   0.15 &  0.29 &  0.00 &  0.00 &  0.10 \\
\textit{WCNN\_25                 }&   0.71 &  0.38 &  0.43 &  1.00 &  1.00 \\
\textit{WCNN\_5                  }&   0.67 &  0.40 &  0.32 &  0.99 &  1.00 \\
\midrule
\textit{GDunn\_d1\_D1            }&   0.58 &  0.44 &  0.01 &  0.74 &  1.00 \\
\textit{GDunn\_d1\_D2            }&   0.59 &  0.45 &  0.01 &  0.77 &  1.00 \\
\textit{GDunn\_d1\_D3            }&   0.59 &  0.45 &  0.01 &  0.77 &  1.00 \\
\textit{GDunn\_d2\_D1            }&   0.40 &  0.34 &  0.07 &  0.37 &  0.63 \\
\textit{GDunn\_d2\_D2            }&   0.46 &  0.32 &  0.14 &  0.48 &  0.70 \\
\textit{GDunn\_d2\_D3            }&   0.46 &  0.33 &  0.13 &  0.46 &  0.72 \\
\textit{GDunn\_d3\_D1            }&   0.40 &  0.35 &  0.02 &  0.38 &  0.63 \\
\textit{GDunn\_d3\_D2            }&   0.43 &  0.36 &  0.08 &  0.44 &  0.79 \\
\textit{GDunn\_d3\_D3            }&   0.42 &  0.36 &  0.07 &  0.41 &  0.77 \\
\textit{GDunn\_d4\_D1            }&   0.40 &  0.36 &  0.01 &  0.38 &  0.63 \\
\textit{GDunn\_d4\_D2            }&   0.42 &  0.36 &  0.07 &  0.43 &  0.73 \\
\textit{GDunn\_d4\_D3            }&   0.41 &  0.36 &  0.03 &  0.39 &  0.72 \\
\textit{GDunn\_d5\_D1            }&   0.13 &  0.20 &  0.02 &  0.07 &  0.14 \\
\textit{GDunn\_d5\_D2            }&   0.16 &  0.19 &  0.02 &  0.09 &  0.23 \\
\textit{GDunn\_d5\_D3            }&   0.12 &  0.28 &  0.00 &  0.00 &  0.01 \\
\bottomrule
\end{tabular}
\end{table}

\begin{table}[p!]
\caption{\label{tab:res2} ARI: Basic summary statistics; part II.
}

\smallskip\centering
\begin{tabular}{lrrrrr}
\toprule
Method                      {} &   Mean &   St.Dev. &    Q1 &   Median &   Q3 \\
\midrule
\textit{DuNN\_5\_Max\_Const      }&   0.47 &  0.44 &  0.09 &  0.26 &  1.00 \\
\textit{DuNN\_5\_Mean\_Const     }&   0.70 &  0.38 &  0.44 &  0.99 &  1.00 \\
\textit{DuNN\_5\_Min\_Const      }&   0.69 &  0.39 &  0.34 &  0.97 &  1.00 \\
\textit{DuNN\_25\_Max\_Const     }&   0.37 &  0.39 &  0.05 &  0.20 &  0.60 \\
\textit{DuNN\_25\_Mean\_Const    }&   0.67 &  0.38 &  0.31 &  0.92 &  1.00 \\
\textit{DuNN\_25\_Min\_Const     }&   0.72 &  0.36 &  0.42 &  0.97 &  1.00 \\
\textit{DuNN\_25\_SMax:5\_Const  }&   0.35 &  0.39 &  0.06 &  0.20 &  0.55 \\
\textit{DuNN\_25\_SMin:5\_Const  }&   0.72 &  0.37 &  0.42 &  0.99 &  1.00 \\
\midrule
\textit{DuNN\_5\_Max\_Min        }&   0.48 &  0.43 &  0.09 &  0.32 &  1.00 \\
\textit{DuNN\_5\_Mean\_Min       }&   0.69 &  0.38 &  0.39 &  0.99 &  1.00 \\
\textit{DuNN\_5\_Min\_Min        }&   0.69 &  0.39 &  0.34 &  0.97 &  1.00 \\
\textit{DuNN\_25\_Max\_Min       }&   0.37 &  0.39 &  0.05 &  0.20 &  0.65 \\
\textit{DuNN\_25\_Mean\_Min      }&   0.67 &  0.38 &  0.31 &  0.92 &  1.00 \\
\textit{DuNN\_25\_Min\_Min       }&   0.72 &  0.35 &  0.42 &  0.97 &  1.00 \\
\textit{DuNN\_25\_SMax:5\_SMin:5 }&   0.36 &  0.38 &  0.05 &  0.24 &  0.39 \\
\textit{DuNN\_25\_SMax:5\_Min    }&   0.34 &  0.39 &  0.01 &  0.20 &  0.48 \\
\midrule
\textit{DuNN\_5\_Max\_Max        }&   0.40 &  0.47 &  0.00 &  0.06 &  1.00 \\
\textit{DuNN\_5\_Mean\_Max       }&   0.57 &  0.44 &  0.06 &  0.61 &  1.00 \\
\textit{DuNN\_5\_Min\_Max        }&   0.68 &  0.40 &  0.31 &  0.97 &  1.00 \\
\textit{DuNN\_25\_Max\_Max       }&   0.38 &  0.41 &  0.00 &  0.19 &  0.89 \\
\textit{DuNN\_25\_Mean\_Max      }&   0.64 &  0.39 &  0.26 &  0.82 &  1.00 \\
\textit{DuNN\_25\_Min\_Max       }&   0.71 &  0.36 &  0.47 &  0.97 &  1.00 \\
\textit{DuNN\_25\_SMin:5\_Max    }&   0.70 &  0.37 &  0.41 &  0.99 &  1.00 \\
\textit{DuNN\_25\_SMin:5\_SMax:5 }&   0.71 &  0.37 &  0.46 &  0.98 &  1.00 \\
\midrule
\textit{DuNN\_5\_Max\_Mean       }&   0.40 &  0.47 &  0.00 &  0.04 &  1.00 \\
\textit{DuNN\_5\_Mean\_Mean      }&   0.60 &  0.44 &  0.04 &  0.93 &  1.00 \\
\textit{DuNN\_5\_Min\_Mean       }&   0.66 &  0.41 &  0.25 &  0.95 &  1.00 \\
\textit{DuNN\_25\_Max\_Mean      }&   0.27 &  0.42 &  0.00 &  0.02 &  0.30 \\
\textit{DuNN\_25\_Mean\_Mean     }&   0.57 &  0.43 &  0.08 &  0.60 &  1.00 \\
\textit{DuNN\_25\_Min\_Mean      }&   0.69 &  0.38 &  0.37 &  0.92 &  1.00 \\
\bottomrule
\end{tabular}
\end{table}

%%%%%%%%%%%%%%%%%%%%%%%%%%%%%%%%%%%%%%%%%%%%%%%%%%%%%%%%%%%%%%%%%%%%%%%%%%%%%%%


\begin{thebibliography}{61}
\expandafter\ifx\csname natexlab\endcsname\relax\def\natexlab#1{#1}\fi
\providecommand{\bibinfo}[2]{#2}
\ifx\xfnm\relax \def\xfnm[#1]{\unskip,\space#1}\fi
%Type = Article
\bibitem[{Aloise et~al.(2009)Aloise, Deshpande, Hansen and
  Popat}]{kmeanseuclidhard}
\bibinfo{author}{D.~Aloise}, \bibinfo{author}{A.~Deshpande},
  \bibinfo{author}{P.~Hansen}, \bibinfo{author}{P.~Popat},
  \bibinfo{title}{{NP}-hardness of {E}uclidean sum-of-squares clustering},
  \bibinfo{journal}{Machine Learning} \bibinfo{volume}{75}
  (\bibinfo{year}{2009}) \bibinfo{pages}{245--248}.
%Type = Article
\bibitem[{Arbelaitz et~al.(2013)Arbelaitz, Gurrutxaga, Muguerza, Pérez and
  Perona}]{ArbelaitzEtAl2013:extensive_CVI}
\bibinfo{author}{O.~Arbelaitz}, \bibinfo{author}{I.~Gurrutxaga},
  \bibinfo{author}{J.~Muguerza}, \bibinfo{author}{J.M. Pérez},
  \bibinfo{author}{I.~Perona}, \bibinfo{title}{An extensive comparative study
  of cluster validity indices}, \bibinfo{journal}{Pattern Recognition}
  \bibinfo{volume}{46} (\bibinfo{year}{2013}) \bibinfo{pages}{243--256}.
%Type = Techreport
\bibitem[{Ball and Hall(1965)}]{BallHall1965:isodata}
\bibinfo{author}{G.~Ball}, \bibinfo{author}{D.~Hall},
  \bibinfo{title}{{ISODATA}: {A} novel method of data analysis and pattern
  classification}, \bibinfo{type}{Technical Report} \bibinfo{number}{AD699616},
  \bibinfo{year}{1965}.
%Type = Manual
\bibitem[{Bendtsen(2012)}]{rpso}
\bibinfo{author}{C.~Bendtsen}, \bibinfo{title}{pso: Particle Swarm
  Optimization}, \bibinfo{year}{2012}. \bibinfo{note}{R package version 1.0.3;
  \url{https://CRAN.R-project.org/package=pso}}.
%Type = Article
\bibitem[{Bezdek et~al.(1984)Bezdek, Ehrlich and Full}]{Bezdek}
\bibinfo{author}{J.~Bezdek}, \bibinfo{author}{R.~Ehrlich},
  \bibinfo{author}{W.~Full}, \bibinfo{title}{{FCM}: {T}he fuzzy c-means
  clustering algorithm}, \bibinfo{journal}{Computers \& Geosciences}
  \bibinfo{volume}{10} (\bibinfo{year}{1984}) \bibinfo{pages}{191--203}.
%Type = Article
\bibitem[{Bezdek and Pal(1998)}]{BezdekPal1998:gdunn}
\bibinfo{author}{J.~Bezdek}, \bibinfo{author}{N.~Pal}, \bibinfo{title}{Some new
  indexes of cluster validity}, \bibinfo{journal}{IEEE Transactions on Systems,
  Man, and Cybernetics, Part B (Cybernetics)} \bibinfo{volume}{28}
  (\bibinfo{year}{1998}) \bibinfo{pages}{301--315}.
%Type = Article
\bibitem[{Brun et~al.(2007)Brun, Sima, Hua, Lowey, Carroll, Suh and
  Dougherty}]{BrunETAL2007:modelcvi}
\bibinfo{author}{M.~Brun}, \bibinfo{author}{C.~Sima}, \bibinfo{author}{J.~Hua},
  \bibinfo{author}{J.~Lowey}, \bibinfo{author}{B.~Carroll},
  \bibinfo{author}{E.~Suh}, \bibinfo{author}{E.R. Dougherty},
  \bibinfo{title}{Model-based evaluation of clustering validation measures},
  \bibinfo{journal}{Pattern Recognition} \bibinfo{volume}{40}
  (\bibinfo{year}{2007}) \bibinfo{pages}{807--824}.
%Type = Article
\bibitem[{Caliński and Harabasz(1974)}]{CalinskiHarabasz1974:index}
\bibinfo{author}{T.~Caliński}, \bibinfo{author}{J.~Harabasz},
  \bibinfo{title}{A dendrite method for cluster analysis},
  \bibinfo{journal}{Communications in {S}tatistics} \bibinfo{volume}{3}
  (\bibinfo{year}{1974}) \bibinfo{pages}{1--27}.
%Type = Phdthesis
\bibitem[{Cena(2018)}]{Cena2018:phd}
\bibinfo{author}{A.~Cena}, \bibinfo{title}{Adaptive hierarchical clustering
  algorithms based on data aggregation methods}, Ph.D. thesis, Systems Research
  Institute, Polish Academy of Sciences, \bibinfo{year}{2018}.
  \bibinfo{note}{In Polish}.
%Type = Article
\bibitem[{Cena and Gagolewski(2020)}]{CenaGagolewski2020:genieowa}
\bibinfo{author}{A.~Cena}, \bibinfo{author}{M.~Gagolewski},
  \bibinfo{title}{{Genie+OWA}: {R}obustifying hierarchical clustering with
  {OWA}-based linkages}, \bibinfo{journal}{Information Sciences}
  \bibinfo{volume}{520} (\bibinfo{year}{2020}) \bibinfo{pages}{324--336}.
%Type = Incollection
\bibitem[{Dasgupta and Ng(2009)}]{multiple_clusterings}
\bibinfo{author}{S.~Dasgupta}, \bibinfo{author}{V.~Ng}, \bibinfo{title}{Single
  data, multiple clusterings}, in: \bibinfo{booktitle}{Proc. NIPS Workshop
  Clustering: Science or Art? Towards Principled Approaches},
  \bibinfo{year}{2009}. \bibinfo{note}{\url{http://clusteringtheory.org}}.
%Type = Article
\bibitem[{Davies and Bouldin(1979)}]{DaviesBouldin1979:index}
\bibinfo{author}{D.L. Davies}, \bibinfo{author}{D.W. Bouldin},
  \bibinfo{title}{A cluster separation measure}, \bibinfo{journal}{IEEE
  Transactions on Pattern Analysis and Machine Intelligence}
  \bibinfo{volume}{PAMI--1} (\bibinfo{year}{1979}) \bibinfo{pages}{224--227}.
%Type = Article
\bibitem[{Dhal et~al.(2021)Dhal, Das, Ray and Gálvez}]{DHAL2021106814}
\bibinfo{author}{K.G. Dhal}, \bibinfo{author}{A.~Das},
  \bibinfo{author}{S.~Ray}, \bibinfo{author}{J.~Gálvez},
  \bibinfo{title}{Randomly attracted rough firefly algorithm for histogram
  based fuzzy image clustering}, \bibinfo{journal}{Knowledge-Based Systems}
  \bibinfo{volume}{216} (\bibinfo{year}{2021}) \bibinfo{pages}{106814}.
%Type = Article
\bibitem[{Dimitriadou et~al.(1999)Dimitriadou, Dolnicar, Leisch and
  Weingessel}]{Dimitriadou1999:more_insights}
\bibinfo{author}{E.~Dimitriadou}, \bibinfo{author}{S.~Dolnicar},
  \bibinfo{author}{F.~Leisch}, \bibinfo{author}{A.~Weingessel},
  \bibinfo{title}{More insight into clustering: {C}omparison of cluster
  algorithms and evaluation of indexes for determining the correct number of
  clusters}, \bibinfo{journal}{Methods of Psychological Research}
  \bibinfo{volume}{4} (\bibinfo{year}{1999}) \bibinfo{pages}{65--66}.
%Type = Article
\bibitem[{Dinh et~al.(2021)Dinh, Huynh and Sriboonchitta}]{DINH2021418}
\bibinfo{author}{D.T. Dinh}, \bibinfo{author}{V.N. Huynh},
  \bibinfo{author}{S.~Sriboonchitta}, \bibinfo{title}{Clustering mixed
  numerical and categorical data with missing values},
  \bibinfo{journal}{Information Sciences} \bibinfo{volume}{571}
  (\bibinfo{year}{2021}) \bibinfo{pages}{418--442}.
%Type = Article
\bibitem[{Du et~al.(2021)Du, Wang, Ji, Wang and Dong}]{DU2021375}
\bibinfo{author}{M.~Du}, \bibinfo{author}{R.~Wang}, \bibinfo{author}{R.~Ji},
  \bibinfo{author}{X.~Wang}, \bibinfo{author}{Y.~Dong}, \bibinfo{title}{{ROBP}
  a robust border-peeling clustering using {C}auchy kernel},
  \bibinfo{journal}{Information Sciences} \bibinfo{volume}{571}
  (\bibinfo{year}{2021}) \bibinfo{pages}{375--400}.
%Type = Misc
\bibitem[{Dua and Graff(2021)}]{uci}
\bibinfo{author}{D.~Dua}, \bibinfo{author}{C.~Graff}, \bibinfo{title}{{UCI}
  {M}achine {L}earning {R}epository}, \bibinfo{year}{2021}.
  \bibinfo{note}{\url{http://archive.ics.uci.edu/ml}}.
%Type = Article
\bibitem[{Dunn(1974)}]{Dunn1974:index}
\bibinfo{author}{J.~Dunn}, \bibinfo{title}{A fuzzy relative of the {ISODATA}
  process and its use in detecting compact well-separated clusters},
  \bibinfo{journal}{Journal of Cybernetics} \bibinfo{volume}{3}
  (\bibinfo{year}{1974}) \bibinfo{pages}{32--57}.
%Type = Article
\bibitem[{Edwards and Cavalli-Sforza(1965)}]{EdwardsSforza1965:divisive}
\bibinfo{author}{A.W.F. Edwards}, \bibinfo{author}{L.L. Cavalli-Sforza},
  \bibinfo{title}{A method for cluster analysis}, \bibinfo{journal}{Biometrics}
  \bibinfo{volume}{21} (\bibinfo{year}{1965}) \bibinfo{pages}{362--375}.
%Type = Article
\bibitem[{Fränti et~al.(2016)Fränti, Mariescu-Istodor and Zhong}]{xnn}
\bibinfo{author}{P.~Fränti}, \bibinfo{author}{R.~Mariescu-Istodor},
  \bibinfo{author}{C.~Zhong}, \bibinfo{title}{{XNN} graph},
  \bibinfo{journal}{Lecture Notes in Computer Science} \bibinfo{volume}{10029}
  (\bibinfo{year}{2016}) \bibinfo{pages}{207--217}.
%Type = Article
\bibitem[{Fränti and Sieranoja(2018)}]{kmsix}
\bibinfo{author}{P.~Fränti}, \bibinfo{author}{S.~Sieranoja},
  \bibinfo{title}{K-means properties on six clustering benchmark datasets},
  \bibinfo{journal}{Applied Intelligence} \bibinfo{volume}{48}
  (\bibinfo{year}{2018}) \bibinfo{pages}{4743--4759}.
%Type = Article
\bibitem[{Gagolewski(2021)}]{genieclust}
\bibinfo{author}{M.~Gagolewski}, \bibinfo{title}{{genieclust}: {F}ast and
  robust hierarchical clustering}, \bibinfo{journal}{SoftwareX}
  \bibinfo{volume}{15} (\bibinfo{year}{2021}) \bibinfo{pages}{100722}.
%Type = Article
\bibitem[{Gagolewski et~al.(2016)Gagolewski, Bartoszuk and
  Cena}]{GagolewskiETAL2016:genie}
\bibinfo{author}{M.~Gagolewski}, \bibinfo{author}{M.~Bartoszuk},
  \bibinfo{author}{A.~Cena}, \bibinfo{title}{Genie: A new, fast, and
  outlier-resistant hierarchical clustering algorithm},
  \bibinfo{journal}{Information Sciences} \bibinfo{volume}{363}
  (\bibinfo{year}{2016}) \bibinfo{pages}{8--23}.
%Type = Misc
\bibitem[{Gagolewski et~al.(2020)}]{clustering_benchmarks_v1}
\bibinfo{author}{M.~Gagolewski}, et~al., \bibinfo{title}{Benchmark suite for
  clustering algorithms -- version 1}, \bibinfo{year}{2020}.
  \bibinfo{note}{\url{https://github.com/gagolews/clustering_benchmarks_v1},
  doi:10.5281/zenodo.3815066}.
%Type = Article
\bibitem[{{Garey} et~al.(1982){Garey}, {Johnson} and
  {Witsenhausen}}]{kmeanshard}
\bibinfo{author}{M.~{Garey}}, \bibinfo{author}{D.~{Johnson}},
  \bibinfo{author}{H.~{Witsenhausen}}, \bibinfo{title}{The complexity of the
  generalized {L}loyd--{M}ax problem}, \bibinfo{journal}{IEEE Transactions on
  Information Theory} \bibinfo{volume}{28} (\bibinfo{year}{1982})
  \bibinfo{pages}{255--256}.
%Type = Article
\bibitem[{Gates and Ahn(2017)}]{ari_adjustment}
\bibinfo{author}{A.J. Gates}, \bibinfo{author}{Y.Y. Ahn}, \bibinfo{title}{The
  impact of random models on clustering similarity}, \bibinfo{journal}{Journal
  of Machine Learning Research} \bibinfo{volume}{18} (\bibinfo{year}{2017})
  \bibinfo{pages}{1--28}.
%Type = Article
\bibitem[{Glover(1986)}]{tabu}
\bibinfo{author}{F.~Glover}, \bibinfo{title}{Future paths for integer
  programming and links to artificial intelligence},
  \bibinfo{journal}{Computers \& Operations Research} \bibinfo{volume}{13}
  (\bibinfo{year}{1986}) \bibinfo{pages}{533--549}.
%Type = Article
\bibitem[{Graves and Pedrycz(2010)}]{GravesPedrycz2010:kernelfuzzyclust}
\bibinfo{author}{D.~Graves}, \bibinfo{author}{W.~Pedrycz},
  \bibinfo{title}{Kernel-based fuzzy clustering: {A} comparative experimental
  study}, \bibinfo{journal}{Fuzzy Sets and Systems} \bibinfo{volume}{161}
  (\bibinfo{year}{2010}) \bibinfo{pages}{522--543}.
%Type = Article
\bibitem[{Halkidi et~al.(2001)Halkidi, Batistakis and
  Vazirgiannis}]{Halkidi2001:cluster_validity}
\bibinfo{author}{M.~Halkidi}, \bibinfo{author}{Y.~Batistakis},
  \bibinfo{author}{M.~Vazirgiannis}, \bibinfo{title}{On clustering validation
  techniques}, \bibinfo{journal}{Journal of Intelligent Information Systems}
  (\bibinfo{year}{2001}) \bibinfo{pages}{107--145}.
%Type = Inproceedings
\bibitem[{{Isimeto} et~al.(2017){Isimeto}, {Yinka-Banjo}, {Uwadia} and
  {Alienyi}}]{8279595}
\bibinfo{author}{R.~{Isimeto}}, \bibinfo{author}{C.~{Yinka-Banjo}},
  \bibinfo{author}{C.O. {Uwadia}}, \bibinfo{author}{D.C. {Alienyi}},
  \bibinfo{title}{An enhanced clustering analysis based on glowworm swarm
  optimization}, in: \bibinfo{booktitle}{2017 IEEE 4th International Conference
  on Soft Computing Machine Intelligence (ISCMI)}, pp. \bibinfo{pages}{42--49}.
%Type = Article
\bibitem[{Jamil and Yang(2013)}]{benchmarkoptim}
\bibinfo{author}{M.~Jamil}, \bibinfo{author}{X.S. Yang}, \bibinfo{title}{A
  literature survey of benchmark functions for global optimization problems},
  \bibinfo{journal}{International Journal of Mathematical Modelling and
  Numerical Optimisation} \bibinfo{volume}{4} (\bibinfo{year}{2013}).
%Type = Article
\bibitem[{Karypis et~al.(1999)Karypis, Han and Kumar}]{chameleon}
\bibinfo{author}{G.~Karypis}, \bibinfo{author}{E.~Han},
  \bibinfo{author}{V.~Kumar}, \bibinfo{title}{{CHAMELEON}: {H}ierarchical
  clustering using dynamic modeling}, \bibinfo{journal}{Computer}
  \bibinfo{volume}{32} (\bibinfo{year}{1999}) \bibinfo{pages}{68--75}.
%Type = Article
\bibitem[{Kim and Ramakrishna(2005)}]{KimRamakrishna2005:CVIassessment}
\bibinfo{author}{M.~Kim}, \bibinfo{author}{R.~Ramakrishna}, \bibinfo{title}{New
  indices for cluster validity assessment}, \bibinfo{journal}{Pattern
  Recognition Letters} \bibinfo{volume}{26} (\bibinfo{year}{2005})
  \bibinfo{pages}{2535--2363}.
%Type = Article
\bibitem[{Kuo et~al.(2021)Kuo, Zheng and Nguyen}]{KUO20211}
\bibinfo{author}{R.~Kuo}, \bibinfo{author}{Y.~Zheng}, \bibinfo{author}{T.P.Q.
  Nguyen}, \bibinfo{title}{Metaheuristic-based possibilistic fuzzy k-modes
  algorithms for categorical data clustering}, \bibinfo{journal}{Information
  Sciences} \bibinfo{volume}{557} (\bibinfo{year}{2021})
  \bibinfo{pages}{1--15}.
%Type = Article
\bibitem[{Lance and Williams(1967)}]{LanceWilliams1967:hierarchicalformula}
\bibinfo{author}{G.~Lance}, \bibinfo{author}{W.~Williams}, \bibinfo{title}{A
  general theory of classification sorting strategies: 1. {H}ierarchical
  systems}, \bibinfo{journal}{Computer Journal}  (\bibinfo{year}{1967})
  \bibinfo{pages}{373--380}.
%Type = Article
\bibitem[{Lawrence and Phipps(1985)}]{HubertArabie1985:partitionscomp}
\bibinfo{author}{H.~Lawrence}, \bibinfo{author}{A.~Phipps},
  \bibinfo{title}{Comparing partitions}, \bibinfo{journal}{Journal of
  Classification} \bibinfo{volume}{2} (\bibinfo{year}{1985})
  \bibinfo{pages}{193--218}.
%Type = Book
\bibitem[{Lee(2011)}]{lee}
\bibinfo{author}{J.~Lee}, \bibinfo{title}{A First Course in Combinatorial
  Optimisation}, \bibinfo{publisher}{Cambridge University Press},
  \bibinfo{year}{2011}.
%Type = Article
\bibitem[{Li et~al.(2016)Li, Zhang, Ding, Zhang and Dale}]{rs8040295}
\bibinfo{author}{H.~Li}, \bibinfo{author}{S.~Zhang}, \bibinfo{author}{X.~Ding},
  \bibinfo{author}{C.~Zhang}, \bibinfo{author}{P.~Dale},
  \bibinfo{title}{Performance evaluation of cluster validity indices (cvis) on
  multi/hyperspectral remote sensing datasets}, \bibinfo{journal}{Remote
  Sensing} \bibinfo{volume}{8} (\bibinfo{year}{2016}).
%Type = Article
\bibitem[{Liang et~al.(2020)Liang, Han and Yang}]{LIANG2020106583}
\bibinfo{author}{S.~Liang}, \bibinfo{author}{D.~Han},
  \bibinfo{author}{Y.~Yang}, \bibinfo{title}{Cluster validity index for
  irregular clustering results}, \bibinfo{journal}{Applied Soft Computing}
  \bibinfo{volume}{95} (\bibinfo{year}{2020}) \bibinfo{pages}{106583}.
%Type = Article
\bibitem[{Liu et~al.(2021)Liu, Jiang, Hou and Liu}]{LIU2021579}
\bibinfo{author}{Y.~Liu}, \bibinfo{author}{Y.~Jiang}, \bibinfo{author}{T.~Hou},
  \bibinfo{author}{F.~Liu}, \bibinfo{title}{A new robust fuzzy clustering
  validity index for imbalanced data sets}, \bibinfo{journal}{Information
  Sciences} \bibinfo{volume}{547} (\bibinfo{year}{2021})
  \bibinfo{pages}{579--591}.
%Type = Article
\bibitem[{Lloyd(1982)}]{lloyd}
\bibinfo{author}{S.~Lloyd}, \bibinfo{title}{Least squares quantization in
  {PCM}}, \bibinfo{journal}{IEEE Transactions on Information Theory}
  \bibinfo{volume}{28} (\bibinfo{year}{1957 (1982)}) \bibinfo{pages}{128--137}.
  \bibinfo{note}{Originally a 1957 Bell Telephone Laboratories Research Report;
  republished in 1982}.
%Type = Article
\bibitem[{{Maulik} and {Bandyopadhyay}(2002)}]{Maulik2002:cvi_comp}
\bibinfo{author}{U.~{Maulik}}, \bibinfo{author}{S.~{Bandyopadhyay}},
  \bibinfo{title}{Performance evaluation of some clustering algorithms and
  validity indices}, \bibinfo{journal}{IEEE Transactions on Pattern Analysis
  and Machine Intelligence} \bibinfo{volume}{24} (\bibinfo{year}{2002})
  \bibinfo{pages}{1650--1654}.
%Type = Article
\bibitem[{Milligan and Cooper(1985)}]{Milligan1985:psycho}
\bibinfo{author}{G.W. Milligan}, \bibinfo{author}{M.C. Cooper},
  \bibinfo{title}{An examination of procedures for determining the number of
  clusters in a data set}, \bibinfo{journal}{Psychometrika}
  \bibinfo{volume}{50} (\bibinfo{year}{1985}) \bibinfo{pages}{159--179}.
%Type = Article
\bibitem[{Mishra et~al.(2021)Mishra, Kar, Mishra, Mohanty and
  Panda}]{MISHRA2021}
\bibinfo{author}{G.~Mishra}, \bibinfo{author}{A.K. Kar}, \bibinfo{author}{A.C.
  Mishra}, \bibinfo{author}{S.K. Mohanty}, \bibinfo{author}{M.~Panda},
  \bibinfo{title}{{SEND}: {A} novel dissimilarity metric using ensemble
  properties of feature space for clustering numerical data},
  \bibinfo{journal}{Information Sciences} \bibinfo{volume}{574}
  (\bibinfo{year}{2021}) \bibinfo{pages}{279--296}.
%Type = Article
\bibitem[{Mullen et~al.(2011)Mullen, Ardia, Gil, Windover and Cline}]{rdeoptim}
\bibinfo{author}{K.~Mullen}, \bibinfo{author}{D.~Ardia},
  \bibinfo{author}{D.~Gil}, \bibinfo{author}{D.~Windover},
  \bibinfo{author}{J.~Cline}, \bibinfo{title}{{DEoptim}: {A}n {R} package for
  global optimization by differential evolution}, \bibinfo{journal}{Journal of
  Statistical Software} \bibinfo{volume}{40} (\bibinfo{year}{2011})
  \bibinfo{pages}{1--26}.
%Type = Incollection
\bibitem[{Müller et~al.(2012)Müller, Nowozin and Lampert}]{itm}
\bibinfo{author}{A.~Müller}, \bibinfo{author}{S.~Nowozin},
  \bibinfo{author}{C.~Lampert}, \bibinfo{title}{Information theoretic
  clustering using minimum spanning trees}, in: \bibinfo{booktitle}{Proc.
  German Conference on Pattern Recognition}, \bibinfo{year}{2012}.
  \bibinfo{note}{\url{https://github.com/amueller/information-theoretic-mst}}.
%Type = Article
\bibitem[{Müllner(2013)}]{Mullner2013:fastcluster}
\bibinfo{author}{D.~Müllner}, \bibinfo{title}{{fastcluster}: Fast
  hierarchical, agglomerative clustering routines for {R} and {{P}ython}},
  \bibinfo{journal}{Journal of Statistical Software} \bibinfo{volume}{53}
  (\bibinfo{year}{2013}) \bibinfo{pages}{1--18}.
%Type = Article
\bibitem[{Nanda and Panda(2014)}]{NANDA20141}
\bibinfo{author}{S.J. Nanda}, \bibinfo{author}{G.~Panda}, \bibinfo{title}{A
  survey on nature inspired metaheuristic algorithms for partitional
  clustering}, \bibinfo{journal}{Swarm and Evolutionary Computation}
  \bibinfo{volume}{16} (\bibinfo{year}{2014}) \bibinfo{pages}{1--18}.
%Type = Book
\bibitem[{Nocedal and Wright(2006)}]{nocedal_wright}
\bibinfo{author}{J.~Nocedal}, \bibinfo{author}{S.J. Wright},
  \bibinfo{title}{Numerical Optimization}, \bibinfo{publisher}{Springer},
  \bibinfo{year}{2006}.
%Type = Article
\bibitem[{Pedregosa et~al.(2011)}]{sklearn}
\bibinfo{author}{F.~Pedregosa}, et~al., \bibinfo{title}{Scikit-learn: {M}achine
  learning in {P}ython}, \bibinfo{journal}{Journal of Machine Learning
  Research} \bibinfo{volume}{12} (\bibinfo{year}{2011})
  \bibinfo{pages}{2825--2830}.
%Type = Book
\bibitem[{Price et~al.(2006)Price, Storn and Lampinen}]{deoptim}
\bibinfo{author}{K.V. Price}, \bibinfo{author}{R.M. Storn},
  \bibinfo{author}{J.A. Lampinen}, \bibinfo{title}{{D}ifferential {E}volution
  -- {A} Practical Approach to Global Optimization},
  \bibinfo{publisher}{Springer-Verlag}, \bibinfo{year}{2006}.
%Type = Article
\bibitem[{Qaddoura et~al.(2020)Qaddoura, Faris and
  Aljarah}]{Qadura2020:nn_ev_cvi}
\bibinfo{author}{R.~Qaddoura}, \bibinfo{author}{H.~Faris},
  \bibinfo{author}{I.~Aljarah}, \bibinfo{title}{An efficient evolutionary
  algorithm with a nearest neighbor search technique for clustering analysis},
  \bibinfo{journal}{Ambient Intell Human Comput}  (\bibinfo{year}{2020}).
%Type = Article
\bibitem[{Rezaei and Fränti(2016)}]{psi}
\bibinfo{author}{M.~Rezaei}, \bibinfo{author}{P.~Fränti}, \bibinfo{title}{Set
  matching measures for external cluster validity}, \bibinfo{journal}{IEEE
  Transactions on Knowledge and Data Engineering} \bibinfo{volume}{28}
  (\bibinfo{year}{2016}) \bibinfo{pages}{2173--2186}.
%Type = Article
\bibitem[{Rousseeuw(1987)}]{Rousseeuw1987:silhouettes}
\bibinfo{author}{P.J. Rousseeuw}, \bibinfo{title}{Silhouettes: {A} graphical
  aid to the interpretation and validation of cluster analysis},
  \bibinfo{journal}{Journal of Computational and Applied Mathematics}
  \bibinfo{volume}{20} (\bibinfo{year}{1987}) \bibinfo{pages}{53--65}.
%Type = Incollection
\bibitem[{Ultsch(2005)}]{fcps}
\bibinfo{author}{A.~Ultsch}, \bibinfo{title}{{C}lustering with {SOM}: {U*C}},
  in: \bibinfo{booktitle}{Workshop on Self-Organizing Maps},
  \bibinfo{publisher}{WSOM 2005}, \bibinfo{year}{2005}, pp.
  \bibinfo{pages}{75--82}.
%Type = Inproceedings
\bibitem[{Vij and Khandnor(2016)}]{VijKhandnor2016:CVIsummary}
\bibinfo{author}{A.~Vij}, \bibinfo{author}{P.~Khandnor},
  \bibinfo{title}{Validity of internal cluster indices}, in:
  \bibinfo{booktitle}{International {C}onference on {C}omputational {S}ystems
  for {S}ustainable {S}olutions}, pp. \bibinfo{pages}{388--395}.
%Type = Article
\bibitem[{{Ward Jr.}(1963)}]{Ward1963:hier}
\bibinfo{author}{J.H. {Ward Jr.}}, \bibinfo{title}{Hierarchical grouping to
  optimize an objective function}, \bibinfo{journal}{Journal of the American
  Statistical Association} \bibinfo{volume}{58} (\bibinfo{year}{1963})
  \bibinfo{pages}{236--244}.
%Type = Article
\bibitem[{Xu et~al.(2020)Xu, Zhang, Liu and Luo}]{XU2020:external_synthetic}
\bibinfo{author}{Q.~Xu}, \bibinfo{author}{Q.~Zhang}, \bibinfo{author}{J.~Liu},
  \bibinfo{author}{B.~Luo}, \bibinfo{title}{Efficient synthetical clustering
  validity indexes for hierarchical clustering}, \bibinfo{journal}{Expert
  Systems with Applications} \bibinfo{volume}{151} (\bibinfo{year}{2020})
  \bibinfo{pages}{113367}.
%Type = Article
\bibitem[{{Xu} et~al.(2012){Xu}, {Xu} and {Wunsch}}]{6170593}
\bibinfo{author}{R.~{Xu}}, \bibinfo{author}{J.~{Xu}}, \bibinfo{author}{D.C.
  {Wunsch}}, \bibinfo{title}{A comparison study of validity indices on
  swarm-intelligence-based clustering}, \bibinfo{journal}{IEEE Transactions on
  Systems, Man, and Cybernetics, Part B (Cybernetics)} \bibinfo{volume}{42}
  (\bibinfo{year}{2012}) \bibinfo{pages}{1243--1256}.
%Type = Article
\bibitem[{Yager(1988)}]{Yager1988:owa}
\bibinfo{author}{R.R. Yager}, \bibinfo{title}{On ordered weighted averaging
  aggregation operators in multicriteria decision making},
  \bibinfo{journal}{IEEE Transactions on Systems, Man, and Cybernetics}
  \bibinfo{volume}{18} (\bibinfo{year}{1988}) \bibinfo{pages}{183--190}.
%Type = Article
\bibitem[{Zhu et~al.(2020)Zhu, Xu and Goodman}]{ZhuEtAl2020:evolution_cvi}
\bibinfo{author}{S.~Zhu}, \bibinfo{author}{L.~Xu}, \bibinfo{author}{E.D.
  Goodman}, \bibinfo{title}{Evolutionary multi-objective automatic clustering
  enhanced with quality metrics and ensemble strategy},
  \bibinfo{journal}{Knowledge-Based Systems} \bibinfo{volume}{188}
  (\bibinfo{year}{2020}) \bibinfo{pages}{105018}.

\end{thebibliography}
\end{document}